\documentclass[nohyperref]{article}

\usepackage{microtype}
\usepackage{graphicx}
\usepackage{subfigure}
\usepackage{adjustbox}
\usepackage{booktabs} 

\usepackage{hyperref}

\usepackage[accepted]{icml2023}

\usepackage{amsmath}
\usepackage{amssymb}
\usepackage{mathtools}
\usepackage{amsthm}
\usepackage{bbm}
\usepackage{multirow}
\usepackage[capitalize,noabbrev]{cleveref}

\theoremstyle{plain}

\theoremstyle{definition}

\theoremstyle{remark}

\usepackage[textsize=tiny]{todonotes}

\icmltitlerunning{Explore and Exploit the Diverse Knowledge in Model Zoo for Domain Generalization}

\begin{document}

\twocolumn[
\icmltitle{Explore and Exploit the Diverse Knowledge in Model Zoo\\ for Domain Generalization}



\icmlsetsymbol{equal}{*}

\begin{icmlauthorlist}
\icmlauthor{Yimeng Chen}{amss,ucas}
\icmlauthor{Tianyang Hu}{noah}
\icmlauthor{Fengwei Zhou}{noah}
\icmlauthor{Zhenguo Li}{noah}
\icmlauthor{Zhiming Ma}{amss,ucas}
\end{icmlauthorlist}

\icmlaffiliation{amss}{Academy of Mathematics and Systems Science, Chinese Academy of Sciences}
\icmlaffiliation{ucas}{University of Chinese Academy of Sciences}
\icmlaffiliation{noah}{Huawei Noah's Ark Lab}

\icmlcorrespondingauthor{Tianyang Hu}{hutianyang1@huawei.com}

\icmlkeywords{Machine Learning, ICML}

\vskip 0.3in
]



\printAffiliationsAndNotice{}  

\begin{abstract}
The proliferation of pretrained models, as a result of advancements in pretraining techniques, has led to the emergence of a vast zoo of publicly available models. Effectively utilizing these resources to obtain models with robust out-of-distribution generalization capabilities for downstream tasks has become a crucial area of research. Previous research has primarily focused on identifying the most powerful models within the model zoo, neglecting to fully leverage the diverse inductive biases contained within. This paper argues that the knowledge contained in weaker models is valuable and presents a method for leveraging the diversity within the model zoo to improve out-of-distribution generalization capabilities. Specifically, we investigate the behaviors of various pretrained models across different domains of downstream tasks by characterizing the variations in their encoded representations in terms of two dimensions: diversity shift and correlation shift. This characterization enables us to propose a new algorithm for integrating diverse pretrained models, not limited to the strongest models, in order to achieve enhanced out-of-distribution generalization performance. Our proposed method demonstrates state-of-the-art empirical results on a variety of datasets, thus validating the benefits of utilizing diverse knowledge.
\end{abstract}

\section{Introduction}

Although remarkable success has been achieved on multiple benchmarks, machine learning models encounter failures in their real-world applications~\cite{volk2019towards,beery2018recognition,dai2018dark}. A central cause for such failures has been recognized as the vulnerability to the \emph{distribution shifts} of the test data~\cite{arjovsky2019invariant,gulrajani2020search}. This can occur when test data is collected under new conditions such as different weather~\cite{volk2019towards}, locations~\cite{beery2018recognition}, or light conditions~\cite{dai2018dark}, resulting in a distribution that differs from the training set.\par

To address this challenge, the task of domain generalization (DG) has gained significant attention, where models are trained on multiple source domains in order to improve their generalizability to unseen domains~\cite{gulrajani2020search}. Multiple DG algorithms have been proposed from various perspectives.
However, this problem is still far from being resolved. For example, Ye et al. (\citeyear{ye2022ood}) have identified two distinct categories of data distribution shifts, namely \emph{diversity shift} and \emph{correlation shift}, and empirically observed that the majority of existing algorithms are only able to surpass the simple empirical risk minimization (ERM) in at most one of the categories. \par

Exploiting pretrained models (PTMs) has shown to be one of the most promising directions for addressing the challenge of DG tasks~\cite{wiles2022a,ye2022ood}. Research has demonstrated that pretraining can provide a significant improvement in performance for DG tasks~\cite{wiles2022a}. 
The growing PTM hubs further bring in great opportunities. With the thriving of pretraining technologies, we now have a huge amount of pretrained models (PTMs) published. For example, Hugging Face Hub~(\citeyear{hughub}) contains over 80K models that vary in data sources, architectures, and pretraining frameworks. Such a zoo of PTMs thus enjoys both high transfer ability and diversity. 
By selecting optimal PTMs for given DG datasets from a zoo of PTMs, Dong et al.~(\citeyear{dong2022zood}) boosted the state-of-the-art DG performance on some benchmarks for over 14\%. \par

While utilizing PTMs has proven to be a promising approach for domain generalization, it remains unclear how to effectively leverage the diverse inductive biases present in different PTMs. Ensemble methods of PTMs have been explored~\cite{dong2022zood,you2021logme}, however, these methods typically only consider the top-performing models based on performance ranking scores. For example, Dong et al.~(\citeyear{dong2022zood}) proposed a feature selection method on the concatenated features of the top-3 ranked PTMs. However, without incorporating diversity, such ensembles can perform worse than single models. 
Although some previous studies have examined certain characteristics of different PTMs~\cite{gontijo-lopes2022no,idrissi2022imagenet}, they are not specified for DG tasks but focus on the in-distribution behavior of the models. This makes it unclear how to effectively utilize these analyses for tackling DG tasks.

To address this challenge, it is crucial to first investigate the compatibility of different PTMs on specific DG tasks and to understand their inductive biases as thoroughly as possible. To achieve this, we propose to profile the shift behaviors of each PTM when conditioned on a given DG task, and then to design an ensemble algorithm that can effectively utilize the profiled shift types.
Specifically, similar to the definition presented in~\cite{ye2022ood}, we interpret the behaviors of PTMs across different domains of downstream tasks by characterizing the variation in their encoded representations from two dimensions, namely \emph{feature diversity shift} and \emph{feature correlation shift}. Through this design, we empirically demonstrate that the differences in shift patterns not only exist among datasets but also among different PTMs.

Such profiling provides guidance for utilizing the inductive bias of poorly performed models which have typical shift patterns on one of the dimensions. As these models capture features that induce a specific kind of distribution shift, we can design ensemble algorithms that prevent the classifier from encountering similar failures, thus improving the out-of-distribution (OOD) generalization ability. \par

To accomplish this, we introduce two key components in our ensemble algorithm: the sample reweight module and the independence penalization module. The sample reweight module utilizes the output of a correlation shift-dominated model to balance the weights of sub-populations, while the independence penalization module requires the main classifier's output to be independent of features that encounter significant diversity shifts among domains. These ensemble procedures are applied during the training process, introducing no additional computational cost for inference.\par

We empirically verify the value of such model zoology on image classification benchmarks, with a model zoo that consists of 35 PTMs varying in architecture, pretraining algorithm, and datasets. The results of our empirical analysis demonstrate the effectiveness of our approach in leveraging poor models to enhance performance, as our new algorithm outperforms top model ensembles. We show that the selected models are different across different datasets, which indicates that our method is adaptive to the specific DG tasks. \par

Our contributions can be summarized as follows.
\begin{itemize}
    \item 
    We propose a novel methodology for profiling the behavior of pretrained models (PTMs) on a given domain generalization (DG) task by quantifying the distribution shift of the features from two dimensions, namely feature diversity shift and feature correlation shift.
    \item 
    We introduce a new ensemble algorithm that leverages the insights from the profiled shift types to effectively utilize the diverse inductive bias among different PTMs for DG tasks.
    \item 
    Through extensive experiments on image classification DG benchmarks, we demonstrate the effectiveness of our proposed approach, which outperforms  top-performing PTM ensembles.
\end{itemize}

This work provides a new perspective on how to effectively leverage the diverse inductive bias of PTMs for domain generalization tasks and highlights the importance of understanding the shift behaviors of models for such tasks.

\section{Related Works}
\paragraph{Domain generalization.}
Numerous domain generalization algorithms have been proposed to alleviate the accuracy degradation caused by distribution shifts via exploiting training domain information ~\cite{arjovsky2019invariant,krueger2021out,Li2018LearningTG,bai2020decaug,Kuang2018,bai2021ood,Cha2021SWADDG,wang2022out,yi2023breaking}. However, \cite{gulrajani2020search} empirically show that recent domain generalization algorithms show no improvement compared with ERM. More fine-grained analyses are further conducted~\cite{ye2022ood,wiles2022a}, where distribution shifts are decomposed into multiple categories. Ye et al.~(\citeyear{ye2022ood}) empirically observed that the majority of the algorithms are only able to surpass the simple ERM in at most one kind of distribution shift. Wiles et al.~(\citeyear{wiles2022a}) show that progress has been made over a standard ERM baseline. Though best methods are not consistent over different data shifts, pretraining and augmentations usually offer large gains.

\paragraph{PTMs for domain generalization.} 
Methods leveraging pretraining models have shown promising improvements in domain generalization performance~\cite{wiles2022a,li2022domain,arpit2021ensemble,dong2022zood,wortsman2022model,rame2022diverse,rame2022recycling}. Among them, ensemble methods combined with PTMs show further advantages. Weight averaging methods combine weights of PTMs of the same architecture over different runs~\cite{rame2022diverse,wortsman2022model} or tasks~\cite{rame2022recycling}. Arpit et al.~(\citeyear{arpit2021ensemble}) ensemble the predictions of moving average models. Recent methods~\cite{li2022domain,dong2022zood} further consider the ensemble of models with different architectures to exploit the growing large PTM hubs. Specifically, Li et al.~(\citeyear{li2022domain}) ensemble predictions of multiple different PTMs via instance-specific attention weights.  ZooD~\cite{dong2022zood} releases the inference cost by only concatenating the representations of top models selected from a diverse model zoo and further conducts Bayesian feature selection. However, as shown in~\cite{dong2022zood}, such an ensemble does not always outperform the single model. The diversity in the model zoo has not been fully understood and exploited, which is the focus of this paper.

\paragraph{Understanding PTMs.} The paradigm of PTM reusing triggers the need for understanding the behavior of a PTM on a given downstream task. Recently, studies on the difference in PTM features have been proposed~\cite{gontijo-lopes2022no,idrissi2022imagenet}, which focus on the in-distribution behavior of models.
Gontijo-Lopes et al.~(\citeyear{gontijo-lopes2022no}) suggest that models under different pretraining techniques learn diverse features. They propose that the correct predictions of high-accuracy models do not dominate those of low-accuracy models, and model ensembles with diverse training methodologies yield the best downstream performance. 
Idrissi et al.~(\citeyear{idrissi2022imagenet}) introduced ImageNet-X, which is a set of human annotations pinpointing failure types for the ImageNet~\cite{russakovsky2015imagenet} dataset. ImageNet-X labels distinguishing object factors (e.g. pose, color) for each image in the validation set and a random subset. They found that most models when trained, fine-tuned, or evaluated on ImageNet, have the same biases. However, this paper shows different observations on the DG datasets, which will be further discussed in Section~\ref{subsec:obs}.

\section{Model Exploration}

To effectively leverage diversity within a model zoo, we need to understand the difference between PTMs conditioned on each specific DG task. To accomplish this, we propose analyzing and describing the changes in PTM feature distributions across downstream domains.

\subsection{Feature Diversity and Correlation Shifts}
\label{subsec:metrics}
Consider a dataset $\mathcal{D}$ that contains samples collected under multiple domains $\mathcal{E}$, i.e., $\mathcal{D} = \{D_e\}_{e\in \mathcal{E}}$. $D_e=\left\{x_{i}^{e}, y_{i}^{e}\right\}_{i=1}^{n^{e}}$ contains instances of random variables $(X, Y)$ that are \emph{i.i.d.} sampled from the probability distribution $\mathbb{P}^e(\mathcal{X} \times \mathcal{Y})$. Consider a PTM that can be viewed as a feature encoder $\phi: \mathcal{X} \rightarrow \mathcal{Z}_{\phi}$. To understand the behavior of such an encoder between different domains, we are in fact concerned with the difference between the distributions of $(\phi(X), Y)$ on different $\mathbb{P}^e, 
\forall e \in \mathcal{E}$. As $\mathbb{P}^e(\phi(X), Y) = \mathbb{P}^e(Y|\phi(X))\mathbb{P}^e(\phi(X))$, the variation of $\mathbb{P}^e(\phi(X), Y)$ can be decomposed into the shift of $\mathbb{P}^e(\phi(X))$ and the shift of $\mathbb{P}^e(Y|\phi(X))$, namely the \emph{feature diversity shift} and the \emph{ feature correlation shift}.

In this paper, we use the following two metrics for measuring the diversity shift and correlation shift of $\phi: \mathbf{x} \mapsto \mathbf{z}$ between a pair of domains $e, e'$, respectively:
\begin{align*}
    F_{div}(\phi, e, e') &= \frac{1}{2} \int_{\mathcal{S}} |p_e(\mathbf{z}) - p_{e'}(\mathbf{z})| \,\mathrm{d}\mathbf{z}, \\
    F_{cor}(\phi, e, e') &= \frac{1}{2} \int_{\mathcal{T}} \tilde{p}_{e, e'}(\mathbf{z})
    \sum_{y \in \mathcal{Y}} |p_{e}(y|\mathbf{z}) - p_{e'}(y|\mathbf{z})| \,\mathrm{d}\mathbf{z},
\end{align*}
where $\tilde{p}_{e,e'}$ is an geometric average of $p_e$ and $p_{e'}$. $\mathcal{S}$ and $\mathcal{T}$ are partitions of the image set $Z_{\phi}$ of $\phi$ defined as follows:
\begin{align*}
    \mathcal{S}(\phi, e, e') &:= \{ \mathbf{z} \in \mathcal{Z}_{\phi} | p_e(\mathbf{z}) \cdot p_{e'}(\mathbf{z}) = 0 \}, \\
    \mathcal{T}(\phi, e, e') &:= \{ \mathbf{z} \in \mathcal{Z}_{\phi} | p_e(\mathbf{z}) \cdot p_{e'}(\mathbf{z}) \neq 0 \}.
\end{align*}
Intuitively, $F_{div}$ describes the proportion of values of features $\phi(\mathbf{x})$ not shared between two domains. 
$F_{cor}$ measures how the correlation between the features and the target label changes between domains. Such definitions are similar to that of diversity shift and correlation shift of datasets in OOD-Bench~\cite{ye2022ood}. Note that the two metrics in this paper are defined for general feature encoders, not a specific encoder $Z_2$ which encodes the latent spurious variable assumed in the data generating process as in~\cite{ye2022ood}. By specific to that encoder, \cite{ye2022ood} view the two metrics as a characteristic of the dataset itself. In contrast, we focus on the difference between general encoders on a given dataset. That generality requires a new design for the estimation methods of the two metrics than that in \cite{ye2022ood}. We further introduce the practical estimation method we proposed in Section~\ref{subsec:estimation}.  \par

\begin{figure}[t]
  \begin{center}
    \includegraphics[scale=0.4]{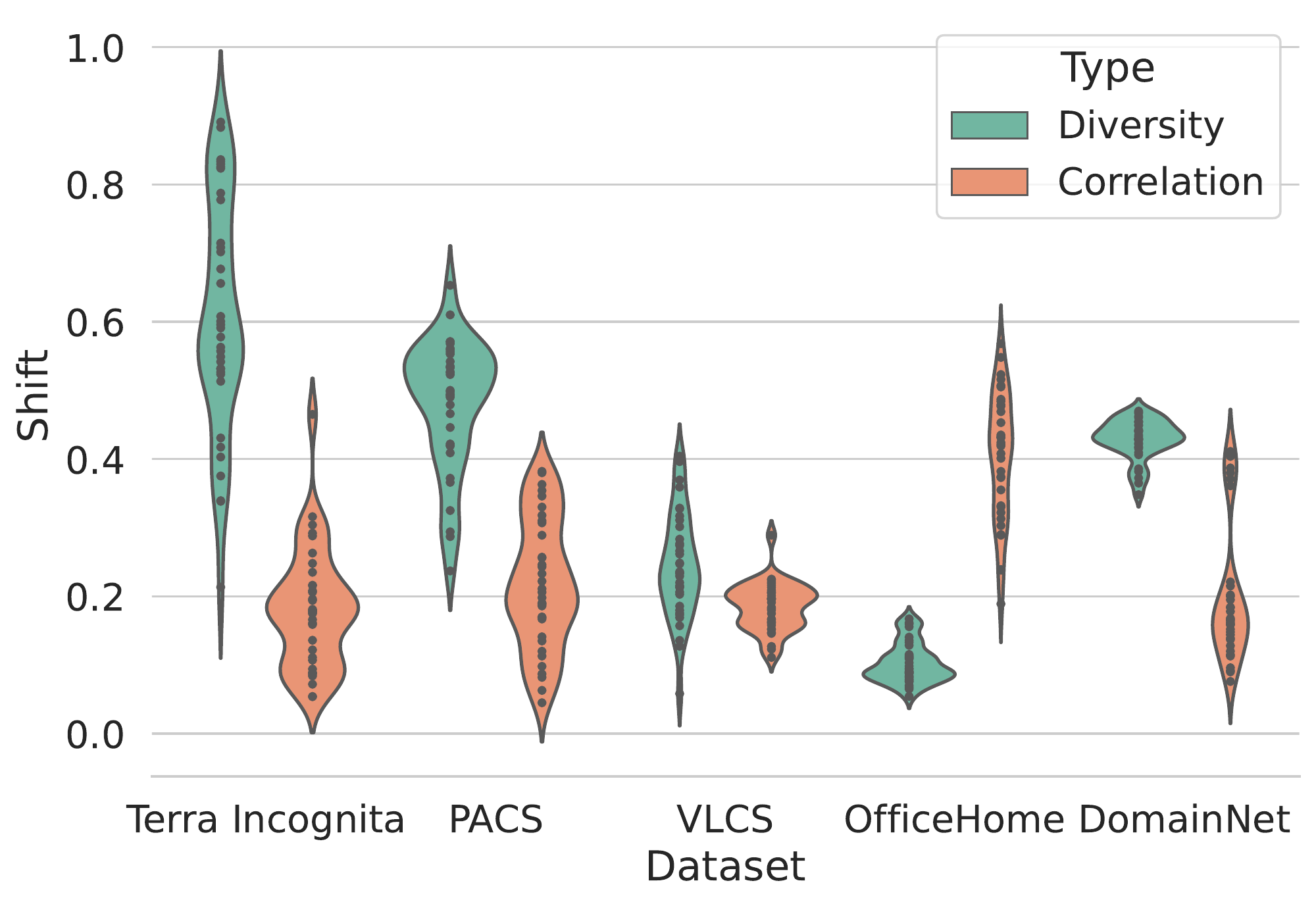}
    \caption{The distribution of feature diversity and correlation shift scores of 35 PTMs on 5 datasets in the DomainBed.}
    \label{fig:dataset}
  \end{center}
\end{figure}

\paragraph{Relation with OOD performance.} For diversity shift, the model's decision on data from the set $\mathcal{S}$ depends on the classification layer's extrapolation behavior, which is hard to infer with in-distribution data. For correlation shift, it directly causes the change of prediction precision and results in the gap between in-distribution and out-of-distribution performance. As a result, we would prefer a representation with both low diversity and correlation shifts so that the in-distribution training controls the out-of-distribution error. Note that by splitting the data into $\mathcal{S}$ and $\mathcal{T}$, we leave out the part that is affected by the classification layer's extrapolation behavior in the correlation shift estimation and the in-domain density shift in the diversity shift estimation. This is the main difference from the scores designed in ZooD.

\subsection{Practical Estimation}
\label{subsec:estimation}
In this section, we show how the two metrics can be computed practically for general latent features of an arbitrary PTM.

\paragraph{Diversity shift.} Denote $\mathcal{S}_{e}(e', \phi) := \{\mathbf{z} \in \mathcal{Z}_{\phi}| p_e(\mathbf{z}) > 0, p_{e'}(\mathbf{z}) = 0 \}$, $\mathcal{S}_{e'}(e, \phi) := \{\mathbf{z} \in \mathcal{Z}_{\phi}| p_{e}(\mathbf{z}) = 0, p_{e'}(\mathbf{z}) > 0 \}$,
$F_{div}(\phi, e, e')$ can be written as
\begin{align*}
 F_{div}(\phi, e, e') = 
    \frac{1}{2} ( \mathbb{P}^e[\mathcal{S}_{e}(e', \phi)] + \mathbb{P}^{e'}[\mathcal{S}_{e'}(e, \phi)]).
\end{align*}
We design the following empirical estimation of $\mathbb{P}^e[\mathcal{S}_{e}(e', \phi)]$:
\begin{align*}
    \hat{\mathbb{P}}^{e}[\hat{\mathcal{S}_{e}}(e', \phi)] := \hat{\mathbb{P}}^{e} (\{ \mathbf{x} \in D_e | \hat{p}_{e'}(\mathbf{z}) < \epsilon_{e'}, \mathbf{z} = \phi(\mathbf{x}) \}).
\end{align*}
Intuitively, we estimate the no-overlap set $\mathcal{S}_e(e', \phi)$ using the estimated probability of the instance in the estimated distribution $\hat{p}_{e'}$. When the probability is lower than a given small threshold $\epsilon_{e'}$, it is considered as in the set $\mathcal{S}_e(e', \phi)$. The threshold $\epsilon_{e'}$ is estimated by
\begin{align*}
    \hat{\mathbb{P}}^{e'} (\{ \mathbf{x} \in V_{e'} | \hat{p}_{e'}(\mathbf{z}) < \epsilon_{e'}, \mathbf{z} = \phi(\mathbf{x}) \}) = 0.01.
\end{align*}
We approximate $p_{e}$ with a Gaussian distribution $\mathcal{N}(\mu_{e}, \Sigma_e)$, and estimate the parameters with empirical statistics on $D_e$. In the same way we can get the estimation of $\mathbb{P}^{e'}[\mathcal{S}_{e'}(e, \phi)]$. The empirical diversity metric is then the average of the two estimations.

\begin{figure*}[t]
  \begin{center}
  \includegraphics[scale=0.45]{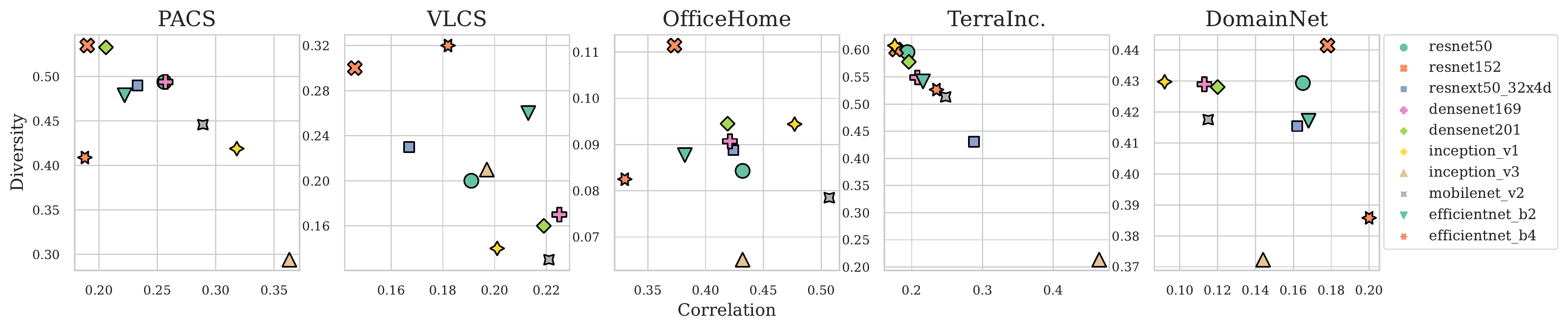}
    \caption{Results of PTMs with different architectures pretrained under the empirical risk minimization framework on ImageNet-1K. Details of these PTMs are provided in Table~\ref{tab_supp:pre-trained_models}.}
    \label{fig:ermcnn}
  \end{center}
\end{figure*}

\begin{figure*}
  \begin{center}
  \includegraphics[scale=0.45]{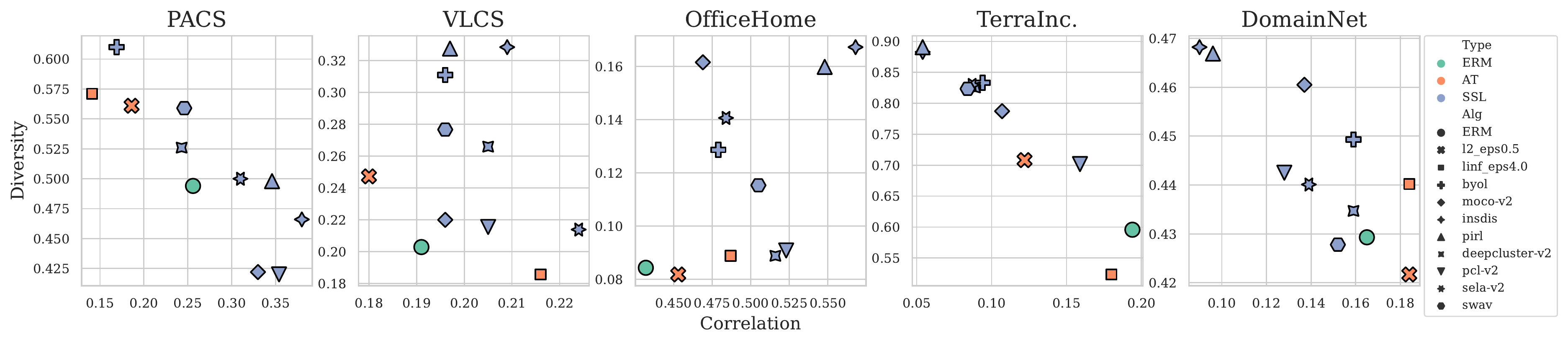}
    \caption{Results comparing ResNet-50s pretrained on ImageNet under different pretraining frameworks. \emph{Type} denotes different pretraining types, including ERM (empirical risk minimization), AT (adversarial training), and SSL (self supervised learning). \emph{Alg} denotes the specific pretraining algorithm.}
    \label{fig:rn50}
  \end{center}
\end{figure*}

\paragraph{Correlation shift.} For each pair of domain $e, e'$. We have the empirical set $\hat{\mathcal{T}}(\phi, e, e') := (D_e \setminus \hat{S}_e(e', \phi)) \cup (D_{e'} \setminus \hat{S}_{e'}(e, \phi)) $. Denote $p_{e, e'} = \frac{1}{2}(p_e + p_{e'})$ and
\begin{align*}
    \hat{D}_{cor} &= \frac{1}{2} \sum_{\mathbf{x} \in \hat{\mathcal{T}}} \hat{p}_{e, e'}(\mathbf{x}) \sum_{y \in \mathcal{Y}} |\hat{p}_{e}(y|\phi(\mathbf{x})) - \hat{p}_{e'}(y|\phi(\mathbf{x}))|.
\end{align*}
As $D_e, D_{e'}$ are independently sampled, $\hat{p}_{e, e'}(\mathbf{x})$ can be estimated by the empirical distribution, i.e., $\hat{p}_{e, e'}(\mathbf{x}) = 1 / |D_e \cup D_e'| $. To estimate $\hat{p}_{e}(y|\phi(\mathbf{x}))$,
we first get a primary estimation $\tilde{p}_{e}(y|\phi(\mathbf{x}))$ with the following equation, where the coefficient matrices $(\mathbf{M}_0, \mathbf{M}_1, \dots,\mathbf{M}_{|\mathcal{Y}|})$ are estimated by minimizing the empirical evidence as in LogME~\cite{you2021logme}, i.e., 
\begin{align*}
    \tilde{p}_{e}(y|\phi(\mathbf{x})) := m(\mathbf{M}_0\phi(\mathbf{x}), \mathbf{M}_1\phi(\mathbf{x}),\dots,\mathbf{M}_{|\mathcal{Y}|}\phi(\mathbf{x})),
\end{align*}
where $m$ denotes the normalization operator. We then calibrate $\tilde{p}_{e}(y|\phi(\mathbf{x}))$ with the empirical accuracy estimated on $\hat{\mathcal{T}}(\phi, e, e')$ to get the final estimation $\hat{p}_{e}(y|\phi(\mathbf{x}))$. More details are provided in Appendix~\ref{apsubsec:estimation}.

\subsection{Observations}
\label{subsec:obs}
In this section, we present the results of our empirical analysis on the distribution shifts of PTMs for different DG datasets. We quantify these shifts using the metrics previously described and discuss the various patterns observed.

We conduct experiments on five domain generalization benchmarks: PACS~\cite{Li2017DeeperBA}, VLCS~\cite{Fang2013UnbiasedML}, Office-Home~\cite{Venkateswara2017DeepHN}, TerraIncognita~\cite{Beery2018RecognitionIT}, DomainNet~\cite{Peng2019MomentMF}. According to \cite{ye2022ood}, PACS, OfficeHome, and TerraIncognita all only encounter diversity shifts, while DomainNet shows both diversity and correlation shifts. 
We adopt the model zoo constructed in~\cite{dong2022zood}, which consists of 35 PTMs with diverse architectures, pre-training methods, and pre-training datasets. The two shift scores for each model are the average of the two metrics in Section~\ref{subsec:metrics} computed on each pair of domains in the dataset. More details are provided in Appendix~\ref{apsubsec:models}. \par

The primary findings in this section are as follows.
\begin{itemize}
    \item Within a specific DG dataset, 
    the shift patterns of PTMs exhibit substantial diversity.
    \item The architectural diversity contributes to distinct shift patterns, and their interrelationships tend to maintain consistency across datasets.
    \item The influence of pretraining frameworks on shift behavior is noteworthy. Particularly, self-supervised learning leads to relatively higher feature diversity shifts.
    \item An increase in the size of the pretraining data results in a decrease in the feature correlation shift.
\end{itemize}
We introduce those findings in detail in the following paragraphs.

\paragraph{Different shift patterns of PTMs on the datasets.} As shown in~\cite{ye2022ood}, different datasets exhibit different trends of shifts. A natural question is how the distribution shift of data interacts with the shift in the feature space of a PTM. The observations in this section show that the shift patterns of PTMs can have a great variety within a given DG dataset. 
Specifically, we compute the average shift metric scores between domain pairs on each dataset. The results are shown in Figure~\ref{fig:dataset}. 
On Terra Incognita, the diversity shift of models varies from 0.21 to 0.89. Notably, some PTMs encounter significant correlation shifts on Terra Incognita, which is different from the dataset correlation shift shown in~\cite{ye2022ood}.

We further compare the results within the following 3 groups of models to show the effect of architectures, training frameworks, and datasets on shift behavior. The details of the 3 groups are introduced in Appendix~\ref{apsubsec:models}.

\paragraph{Architectures.} We compare models with different architectures but pre-trained with the same framework on the same dataset. As shown in Figure~\ref{fig:ermcnn}, when comparing PTMs pretrained under the ERM framework on ImageNet-1K~\cite{russakovsky2015imagenet}, we found that the variation of architectures resulted in a wide range of shift patterns. It can be observed that across different datasets, ResNet-152 generally exhibits a larger diversity shift compared to ResNet-50, and a smaller correlation shift. Additionally, after fine-tuning, ResNet-152 achieves higher OOD accuracy than ResNet-50. These findings suggest an interesting observation that while ResNet-152 captures domain-specific features, they do not result in a geometric skew~\cite{nagarajan2021understanding}.

\paragraph{Pretraining frameworks.} To show the effect of pretraining frameworks, we compare models with a fixed architecture but trained with different optimization objectives on the same dataset. Figure~\ref{fig:rn50} shows the results comparing ResNet-50s pretrained on ImageNet under different frameworks, i.e., ERM, self-supervised learning (SSL), and adversarial training (AT)~\cite{Madry2018TowardsDL}. We can find models pretrained using SSL methods exhibit overall higher diversity shifts. 
This is not unexpected, as SSL methods tend to learn features that maximally preserve the original information of the raw input, including the domain-specific part. For example, generative-based SSL such as the Masked autoencoder~\cite{he2021masked} learns to reconstruct images with only a small fraction of the pixels. Additionally, contrastive learning methods have been observed to suffer from the negative transfer phenomenon~\cite{liu2022task}, where the learned features perform poorly on downstream tasks. Furthermore, the use of cosine similarity in contrastive learning has been noted to result in overly complex feature maps~\cite{hu2022your}, which can negatively impact out-of-distribution generalization.
Among SSL methods, PIRL~\cite{Misra2020SelfSupervisedLO} and InsDis~\cite{Wu2018UnsupervisedFL} usually have the most significant diversity shifts and worse OOD performance on these datasets~\cite{dong2022zood}.

\begin{figure}[t]
  \begin{center}
  \includegraphics[scale=0.5]{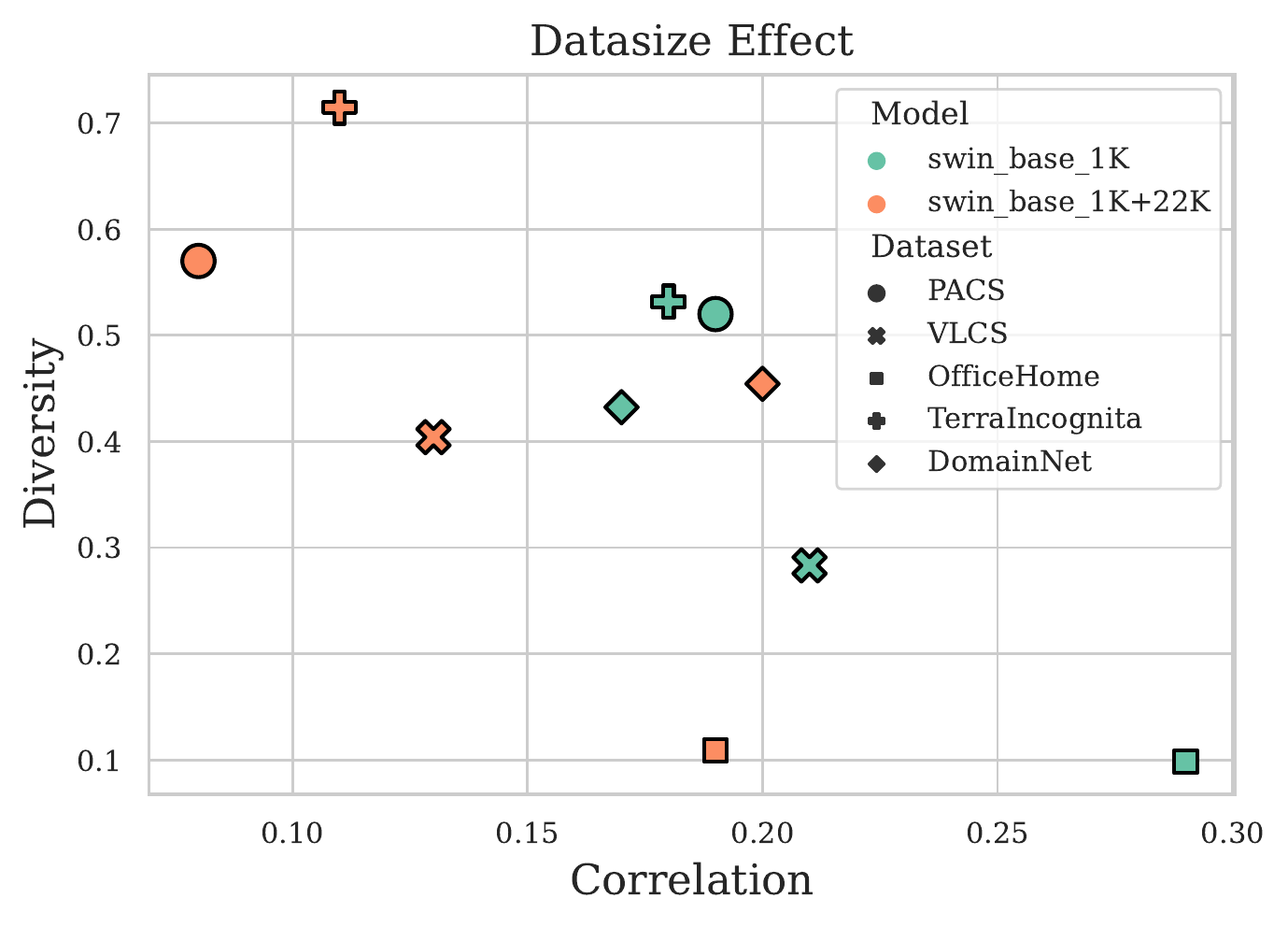}
    \caption{Results of Swin transformers~\cite{Liu2021SwinTH} pretrained on ImageNet-1K and both ImageNet-1K and ImageNet-22K on 5 datasets.}
    \label{fig:swin}
  \end{center}
\end{figure}

\paragraph{Datasets.} To demonstrate the impact of dataset size on the distribution shifts of PTMs, we compare the performance of Swin transformers~\cite{Liu2021SwinTH} pretrained on ImageNet-1K and both ImageNet-1K and ImageNet-22K~\cite{Russakovsky2015ImageNetLS}, as shown in Figure~\ref{fig:swin}. It indicates that the use of larger pretraining data results in a significant decrease in correlation shift, which may be attributed to the increased complexity of the supervised pretraining tasks.

\section{Model Zoo Exploitation}
In this section, we demonstrate how the characteristic of diversity in models can be employed to enhance the domain generalization performance of strong models. In the previous section, we established that models exhibit two distinct types of shift patterns. Our observations indicate that some PTMs are dominated by one type of shift, for example, PIRL on TerraIncognita. This insight inspires the design of an ensemble algorithm that addresses the two dimensions of feature shifts. By leveraging two auxiliary models that are dominated by the two shifts respectively, we design corresponding algorithms to resolve the specific shifts.

\begin{table*}[t]
    \centering
    \caption{Comparison of test domain accuracy between our method and SOTA OOD methods. The results of SWAD are from~\cite{Cha2021SWADDG}, and results denoted with ${\dagger}$ are from~\cite{gulrajani2020search}. The results of three versions of ZooD are from~\cite{dong2022zood} (denoted with $*$). Our results are average of three trials.}
    \label{tab:sota_results}
    \begin{tabular}{l|ccccc|c}
        \toprule
        \textbf{Method} &  \textbf{PACS}  & \textbf{VLCS}  & \textbf{OfficeHome} & \textbf{TerraInc.} & \textbf{DomainNet}   & \textbf{Avg}  \\
        \hline
        ERM$^{\dagger}$           & 85.5 & 77.5 & 66.5 & 46.1 & 40.9 & 63.3 \\ 
        IRM$^{\dagger}$           & 83.5 & 78.6 & 64.3 & 47.6 & 33.9 & 61.6 \\ 
        GroupDRO$^{\dagger}$      & 84.4 & 76.7 & 66.0 & 43.2 & 33.3 & 60.7 \\ 
        I-Mixup$^{\dagger}$       & 84.6 & 77.4 & 68.1 & 47.9 & 39.2 & 63.4 \\ 
        MMD$^{\dagger}$           & 84.7 & 77.5 & 66.4 & 42.2 & 23.4 & 58.8 \\ 
        SagNet$^{\dagger}$        & 86.3 & 77.8 & 68.1 & 48.6 & 40.3 & 64.2 \\ 
        ARM$^{\dagger}$           & 85.1 & 77.6 & 64.8 & 45.5 & 35.5 & 61.7 \\ 
        VREx$^{\dagger}$     & 84.9 & 78.3 & 66.4 & 46.4 & 33.6 & 61.9\\ 
        RSC$^{\dagger}$           & 85.2 & 77.1 & 65.5 & 46.6 & 38.9 & 62.7 \\ 
        SWAD & 88.1 & 79.1 & 70.6 & 50.0 & 46.5 & 66.9 \\
        \hline
        \multicolumn{7}{c}{ZooD} \\
        \hline
        Single$^*$   & 96.0 & 79.5 & 84.6 & 37.3 & 48.2 & 69.1 \\
        Ensemble$^*$ & 95.5 & 80.1 & 85.0 & 38.2 & 50.5 & 69.9 \\
        F. Selection$^*$ & 96.3 & 80.6 & 85.1 & 42.3 & \textbf{50.6} & 71.0 \\
        \hline
        \multicolumn{7}{c}{Ours} \\
        \hline
        Single + Rew & 96.3 & 81.2 & 84.0 & 52.0 & 48.2 & 72.3 \\
        + HSIC & \textbf{96.7} & \textbf{81.5} & 85.2 & 52.3 & 49.2 & 72.8 \\
        + Both & \textbf{96.7} & 81.4 & \textbf{85.3} & \textbf{53.0} & 49.2 & \textbf{73.1} \\          
         \bottomrule
    \end{tabular}
\end{table*}

\subsection{Diversity Ensemble Method}

To prevent potential failure caused by the diversity shift, we utilize the auxiliary model which encodes features that encounter significant diversity shifts. We propose to require the prediction of the main model to be independent of those features thus mitigating the effect of diversity shift on the predictor. To constraint the independence, we adopt a differentiable independence measure, the
Hilbert-Schmidt independence criterion (HSIC)~\cite{gretton2007kernel}.
The idea of using HSIC is inspired by the algorithm proposed in~\cite{bahng2020learning}, where HSIC is used for penalizing the dependency between the predicts of the main model and multiple biased models. \par

Formally, denote $Z_l = l_m \circ f_M(X)$, where $l_m : \mathcal{Z}_M \rightarrow \mathcal{Z}_l $ is the classifier on the top of the main model $f_M: \mathcal{X} \rightarrow \mathcal{Z}_M$. Denote $Z_d = f_d(X)$, where $f_d : \mathcal{X} \rightarrow \mathcal{Z}_d$ is the diversity auxiliary model. Our target is then to constrain the dependency between $Z_l$ and $Z_d$. Denote $k$ as a kernel function on $\mathcal{Z}_d \times \mathcal{Z}_d$, $l$ as a kernel function on $\mathcal{Z}_l \times \mathcal{Z}_l$.
The HSIC statistic between the main model $f_M$ and the auxiliary model $f_d$ is defined as follows:
\begin{align*}
    \textrm{HSIC}(f_M, f_d) := & \mathbb{E} \left[k\left(Z_d, Z_d^{\prime}\right) l\left(Z_l, Z_l^{\prime}\right)\right] + \\
    & \quad \mathbb{E} \left[k\left(Z_d, Z_d^{\prime}\right)\right] \mathbb{E}\left[l\left(Z_l, Z_l^{\prime}\right)\right] \\
& \;\; -2 \mathbb{E}\left[\mathbb{E}_{Z_d^{\prime}}\left[k\left(Z_d, Z_d^{\prime}\right)\right] \mathbb{E}_{Z_l^{\prime}}\left[l\left(Z_l, Z_l^{\prime}\right)\right]\right].
\end{align*}

Instead of the unbiased estimator in~\cite{bahng2020learning}, we used the biased empirical estimate $\mathrm{HSIC}_b$~\cite{gretton2007kernel}: 
\begin{align*}
    \textrm{HSIC}_b(f_M, f_d):= \frac{1}{m^2} \operatorname{trace}(\mathbf{K H L H}), 
\end{align*}
where we suppose the sample size is $m$, $\mathbf{K}$ denotes the 
$m \times m$ matrix with entries $k_{ij} := k(f_d(x_i), f_d(x_j))$, $\mathbf{L}$ denotes the $m \times m$ matrix with entries $l_{ij} := l(l_m \circ f_M(x_i), l_m \circ f_M(x_j))$. $\mathbf{H} = \mathbf{I} - \frac{1}{m}\mathbf{1}\mathbf{1}^T$, where $\mathbf{1}$ is an $m \times 1$ vector of ones.

The final training objective of the main model writes as follows:
\begin{align*}
     \mathcal{L}(f_M) := \min_{f_M} \mathbb{E}_{X, Y \sim \mathbb{P}_\mathcal{D}} & [\mathcal{L}_c (Y, f_M(X)) \\ \nonumber
     & + \lambda \mathrm{HSIC}_d(f_M, f_d)].
\end{align*}
In our implementation, we use the Gaussian kernel $l(z, z') = \exp(-\gamma_1 \lVert z - z' \rVert^2)$, $k(z, z') = \exp(-\gamma_2 \lVert z - z' \rVert^2)$. To mitigate the effect of the dimension, we rescale $\gamma_1$ and $\gamma_2$ by dividing by the dimension of the representation $z$ in the calculation. Following methods in invariant learning literature~\citep{chen2022does}, we introduce an additional hyperparameter $N_{\textrm{warm-up}}$ which controls the number of warm-up steps before the HSIC penalty is added to the loss.

\begin{table*}[!t]
    \centering
    \caption{Main and auxiliary models. HSIC aux. denotes the auxiliary models that are dominated by the diversity shift and adopted for computing the HSIC constraint. Rew. aux. denotes the auxiliary models that are dominated by the correlation shift. The metric values are averaged over each pair of domains in the dataset. Details of the model configuration are provided in Appendix~\ref{apsubsec:config}.}
    \label{tab:models}
    \begin{tabular}{l|ccccc}
        \toprule
        \textbf{Datasets} &  \textbf{PACS}  & \textbf{VLCS}  & \textbf{OfficeHome} & \textbf{TerraInc.} & \textbf{DomainNet} \\
        \midrule
        Main model & CLIP-ViT & CLIP-ViT & Swin-B-22 & Swin-B-22 & ResNext-101 \\
        $F_{div}$ & 0.37 & 0.36 & 0.11 & 0.71 & 0.35 \\ 
        $F_{cor}$ & 0.05 & 0.11 & 0.19 & 0.11 & 0.41 \\
        \midrule
        HSIC aux. & ResNet50-ss & ResNet50-InsDis & ResNet50-InsDis & ResNet50-PIRL & ViT-B \\
        $F_{div}$ & 0.65 & 0.33 & 0.17 & 0.89 & 0.47 \\
        $F_{cor}$ & 0.10 & 0.21 & 0.57 & 0.05 & 0.22 \\
        \midrule
        Rew. aux. & BEiT-base & BEiT-base & deepcluster-v2 & inception-v3 & ResNet50-sws \\
        $F_{div}$ & 0.33 & 0.06 & 0.09 & 0.21 & 0.45 \\
        $F_{cor}$ & 0.38 & 0.29 & 0.52 & 0.47 & 0.40 \\
        \bottomrule
    \end{tabular}
\end{table*}

\subsection{Correlation Ensemble Method}
To prevent potential failure caused by the correlation shift, we adopt the auxiliary model which encodes features that encounter significant correlation shifts. In this module, we reweight training instances to weaken the correlation between the features and the target labels. By that, we avoid the predictor from skewing to that unstable correlation across domains. 

Specifically, denote the auxiliary model as $f_c$ and its uncertainty output for instance $\mathbf{x}$ as $\mathbf{p}_c(\mathbf{x})$.
We follow the classical strategy which has been proven effective in the debias literature~\cite{xiong2021uncertainty} to reweight the instance loss with 
\[w_c(\mathbf{x}, y) = p(y)/p_c(\mathbf{x})_y,\]
where $p_c(\mathbf{x})_y$ is the $y$-th component of $\mathbf{p}_c(\mathbf{x})$. During training steps, the weights in each batch are smoothed with a hyperparameter $T$ and normalized~\citep{yi2021improved}. The loss on a batch $|\mathcal{B}|$ is then
\begin{align*}
    \mathcal{L}_{\mathcal{B}}(f_M) := \frac{1}{|\mathcal{B}|} \sum_{(\mathbf{x}, y) \in \mathcal{D}} m(\frac{p(y)}{p_c(\mathbf{x})_y \cdot T}) \mathcal{L}_c (y, f_M(\mathbf{x})), 
\end{align*}
where $m$ denotes the normalization operation over samples in $\mathcal{B}$. We introduce an additional hyperparameter $N_{\textrm{anneal}}$ which controls the number of annealing steps where $T$ is infinitely large, i.e., before the adjusted weights are attached to the samples.

\section{Experiments}
We conduct experiments on domain generalization benchmarks to evaluate the effectiveness of our proposed zoo exploiting method. Our results demonstrate that it consistently outperforms single top models and improves the performance of top model ensembles, highlighting the benefits of exploiting model diversity. Additionally, we analyze the correlation between OOD accuracy and the feature diversity and correlation shifts of the fine-tuned classifiers.

\subsection{Experiment Settings}
\label{subsec:set}
\paragraph{Datasets.} We conduct experiments on five domain generalization benchmarks:
PACS~\cite{Li2017DeeperBA}, VLCS~\cite{Fang2013UnbiasedML}, OfficeHome~\cite{Venkateswara2017DeepHN} , TerraIncognita~\cite{Beery2018RecognitionIT}, and DomainNet~\cite{Peng2019MomentMF}.
During training on each dataset, one of the domains is chosen as the target domain and the remaining are the training domains, where 20\% samples are used for validation and model selection. The final test accuracy on the dataset is the mean of the test results on each target domain.

\paragraph{Baselines.} We compare the proposed algorithm with previous SOTA OOD methods and three versions of ZooD, including 1) \emph{Single}: fine-tune the top-1 model ranked by ZooD; 2) \emph{Ensemble}:  fine-tune an ensemble of the top-K models; 3) \emph{F. Selection}: fine-tune an ensemble of the top-K models with feature selection, which is the expected result using ZooD. Our algorithm also has three versions. 1) \textbf{Single+Rew}: fine-tune the top-1 model ranked by ZooD with reweight auxiliary; 2) \textbf{Single+HSIC}: fine-tune the top-1 model with HSIC auxiliary; 3) \textbf{Single+Both}: fine-tune the top-1 model with both kinds of auxiliary.

\paragraph{Configurations.} We follow the setting of ZooD to construct a model zoo consisting of 35 PTMs. As discussed in Section~\ref{subsec:obs}, these models vary in architectures, pretraining methods, and datasets. For auxiliary models, we select models that are extreme at one shift metric. For the main model, we use the Top-1 model ranked by ZooD. The detailed statistics of selected auxiliary models and the main models are shown in Table~\ref{tab:models}. We use a 3-layer MLP as the prediction head on the top of the main model and fine-tune it on the downstream tasks. Following ZooD, we adopt the leave-one-domain-out cross-validation setup in DomainBed for hyper-parameter selection and run 3 trials. More details on the experimental setup are in Appendix~\ref{apsubsec:config}.

\begin{table}[t]
    \centering
    \caption{Results on the ensemble of Top-3 models. Results denoted with $*$ are from~\cite{dong2022zood}.}
    \label{tab:ens}
    \begin{tabular}{l|cc}
        \toprule
        \textbf{Datasets} & \textbf{OfficeHome} & \textbf{DomainNet} \\
        \midrule
        Ensemble$^*$ & 85.0 & 50.5 \\
        F. Selection$^*$ & 85.1 & 50.6 \\
        \midrule
        Ensemble+Rew & 85.1 & 50.6 \\
        +HSIC & \textbf{86.0} & \textbf{51.4} \\
        \bottomrule
    \end{tabular}
\end{table}

\begin{table}
    \centering
    \caption{Shift scores of the last layer representation of main predictors. ERM, HSIC, Rew. and Two denote the scores for the logits of the main predictor.}
    \label{tab:logits}
    \begin{tabular}{cl|cc}
        \toprule
        & \textbf{Datasets}& \textbf{VLCS} & \textbf{TerraInc.} \\
        \midrule
        & Model & CLIP-ViT & Swin-B-22 \\
        & Test Env. & 1 & 0 \\
        \midrule
        ERM & $F_{div}$ & 0.11 & 0.34 \\
        & $F_{cor}$ & 0.06 & 0.16 \\
        \midrule
        HSIC & $F_{div}$ & 0.095 & 0.27 \\
        & $F_{cor}$ & 0.041 & 0.17 \\
        \midrule
        Rew. & $F_{div}$ & 0.101 & 0.33 \\
        & $F_{cor}$ & 0.054 & 0.15 \\
        \midrule
        Two & $F_{div}$ & 0.098 & 0.28 \\
        & $F_{cor}$ & 0.050 & 0.16 \\
        \bottomrule
    \end{tabular}
\end{table}

\subsection{Experiment Results}

Table~\ref{tab:sota_results} presents the main results of our proposed methods on the five datasets of the DomainBed benchmark. The results indicate that the incorporation of the independence penalization module and the combination of reweight and independence penalization modules consistently improve the performance of the single top model. On average, the combination of both methods (Single+Both) results in an approximate 6\% improvement in accuracy. Notably, on the PACS, VLCS, and TerraIncognita datasets, our proposed methods even outperform the F. Selection method. This observation highlights the potential of utilizing model inductive bias to leverage weak models in boosting performance, rather than relying solely on strong models.

On the OfficeHome and DomainNet datasets, the proposed methods do not show significant improvements over top-3 ensembles (the Ensemble version of ZooD). To further investigate this, we also conducted experiments using our methods on top-3 ensembles. The results, presented in Table~\ref{tab:ens}, reveal that compared to the F. Selection method, the incorporation of the independence penalization module can significantly enhance the overall accuracy.

It is worth noting that, unlike the independence penalization module, the improvements brought by the reweight module are only significant on the VLCS and TerraIncognita datasets. For the PACS dataset, this may be attributed to the fact that the $F_{cor}$ of the main model is already non-significant, as reported in Table~\ref{tab:models}. For the OfficeHome and DomainNet datasets, this may be due to the limited effectiveness of the reweighting strategy when the number of classes is large (65 and 345). Previous literature has only validated its success on tasks with a number of classes lower than 10~\cite{xiong2021uncertainty}.

To further interpret the results, we analyze the shift pattern of the main predictor. Table~\ref{tab:logits} shows the scores comparison of the last layer features (logits) of the main predictor. The results are obtained using the following hyperparameter set: $\lambda=100, N_{\textrm{warm-up}}=500, \gamma_1=0.5, \gamma_2=0.25$, $T=1, N_{\textrm{anneal}}=2000$. As expected, compared to the results obtained using ERM, HSIC, and Rew. lead to a decrease in $F_{div}$ and $F_{cor}$, respectively. The results obtained using both modules show a compromise between the two modules. It is worth noting that the use of HSIC on the VLCS dataset leads to a significant decrease in $F_{cor}$, which can explain the result in Table~\ref{tab:sota_results} where incorporating the reweight module in Two does not further improve the results of HSIC.

\section{Conclusion}

In this work, we have presented a novel approach for utilizing the diverse knowledge present in a model zoo for domain generalization tasks. 
The main takeaway findings of this study are two-fold. Firstly, it emphasizes that even the most powerful models have the potential for further enhancements in downstream DG tasks. Secondly, it illustrates that the enhancements do not solely come from powerful models, but rather from a combination of models with diverse characteristics, a weak model can also contribute to the enhancement of an already strong model. This highlights the importance of maintaining a diverse zoo of pretrained models for the community.
It is worth emphasizing that our proposed profiling method is general and can be applied to other tasks and domains, making it an interesting avenue for further research. Overall, this work provides a new perspective on how to better utilize the diverse knowledge in a model zoo and opens up new possibilities for improving performance on out-of-distribution tasks.

\bibliography{Reference}

\begin{thebibliography}{69}
\providecommand{\natexlab}[1]{#1}
\providecommand{\url}[1]{\texttt{#1}}
\expandafter\ifx\csname urlstyle\endcsname\relax
  \providecommand{\doi}[1]{doi: #1}\else
  \providecommand{\doi}{doi: \begingroup \urlstyle{rm}\Url}\fi

\bibitem[Arjovsky et~al.(2019)Arjovsky, Bottou, Gulrajani, and
  Lopez-Paz]{arjovsky2019invariant}
Arjovsky, M., Bottou, L., Gulrajani, I., and Lopez-Paz, D.
\newblock Invariant risk minimization.
\newblock \emph{arXiv preprint arXiv:1907.02893}, 2019.

\bibitem[Arpit et~al.(2021)Arpit, Wang, Zhou, and Xiong]{arpit2021ensemble}
Arpit, D., Wang, H., Zhou, Y., and Xiong, C.
\newblock Ensemble of averages: Improving model selection and boosting
  performance in domain generalization.
\newblock \emph{arXiv preprint arXiv:2110.10832}, 2021.

\bibitem[Asano et~al.(2020)Asano, Rupprecht, and
  Vedaldi]{Asano2020SelflabellingVS}
Asano, Y.~M., Rupprecht, C., and Vedaldi, A.
\newblock Self-labelling via simultaneous clustering and representation
  learning.
\newblock \emph{ArXiv}, abs/1911.05371, 2020.

\bibitem[Bahng et~al.(2020)Bahng, Chun, Yun, Choo, and Oh]{bahng2020learning}
Bahng, H., Chun, S., Yun, S., Choo, J., and Oh, S.~J.
\newblock Learning de-biased representations with biased representations.
\newblock In \emph{International Conference on Machine Learning}, pp.\
  528--539. PMLR, 2020.

\bibitem[Bai et~al.(2021{\natexlab{a}})Bai, Sun, Hong, Zhou, Ye, Ye, Chan, and
  Li]{bai2020decaug}
Bai, H., Sun, R., Hong, L., Zhou, F., Ye, N., Ye, H.-J., Chan, S.-H.~G., and
  Li, Z.
\newblock Decaug: Out-of-distribution generalization via decomposed feature
  representation and semantic augmentation.
\newblock In \emph{Proceedings of the AAAI Conference on Artificial
  Intelligence}, volume~35, pp.\  6705--6713, 2021{\natexlab{a}}.

\bibitem[Bai et~al.(2021{\natexlab{b}})Bai, Zhou, Hong, Ye, Chan, and
  Li]{bai2021ood}
Bai, H., Zhou, F., Hong, L., Ye, N., Chan, S.-H.~G., and Li, Z.
\newblock Nas-ood: Neural architecture search for out-of-distribution
  generalization.
\newblock In \emph{Proceedings of the IEEE/CVF International Conference on
  Computer Vision}, pp.\  8320--8329, 2021{\natexlab{b}}.

\bibitem[Bao et~al.(2021)Bao, Dong, and Wei]{bao2021beit}
Bao, H., Dong, L., and Wei, F.
\newblock Beit: Bert pre-training of image transformers.
\newblock \emph{arXiv preprint arXiv:2106.08254}, 2021.

\bibitem[Beery et~al.(2018{\natexlab{a}})Beery, Horn, and
  Perona]{Beery2018RecognitionIT}
Beery, S., Horn, G.~V., and Perona, P.
\newblock Recognition in terra incognita.
\newblock In \emph{ECCV}, 2018{\natexlab{a}}.

\bibitem[Beery et~al.(2018{\natexlab{b}})Beery, Van~Horn, and
  Perona]{beery2018recognition}
Beery, S., Van~Horn, G., and Perona, P.
\newblock Recognition in terra incognita.
\newblock In \emph{Proceedings of the European conference on computer vision
  (ECCV)}, pp.\  456--473, 2018{\natexlab{b}}.

\bibitem[Caron et~al.(2018)Caron, Bojanowski, Joulin, and
  Douze]{Caron2018DeepCF}
Caron, M., Bojanowski, P., Joulin, A., and Douze, M.
\newblock Deep clustering for unsupervised learning of visual features.
\newblock In \emph{Proceedings of the European Conference on Computer Vision
  (ECCV)}, pp.\  132--149, 2018.

\bibitem[Caron et~al.(2020)Caron, Misra, Mairal, Goyal, Bojanowski, and
  Joulin]{Caron2020UnsupervisedLO}
Caron, M., Misra, I., Mairal, J., Goyal, P., Bojanowski, P., and Joulin, A.
\newblock Unsupervised learning of visual features by contrasting cluster
  assignments.
\newblock In \emph{Thirty-fourth Conference on Neural Information Processing
  Systems (NeurIPS)}, 2020.

\bibitem[Cha et~al.(2021)Cha, Chun, Lee, Cho, Park, Lee, and
  Park]{Cha2021SWADDG}
Cha, J., Chun, S., Lee, K., Cho, H.-C., Park, S., Lee, Y., and Park, S.
\newblock Swad: Domain generalization by seeking flat minima.
\newblock \emph{arXiv preprint arXiv:2102.08604}, 2021.

\bibitem[Chen et~al.(2020)Chen, Fan, Girshick, and He]{Chen2020ImprovedBW}
Chen, X., Fan, H., Girshick, R.~B., and He, K.
\newblock Improved baselines with momentum contrastive learning.
\newblock \emph{ArXiv}, abs/2003.04297, 2020.

\bibitem[Chen et~al.(2022)Chen, Xiong, Ma, and Lan]{chen2022does}
Chen, Y., Xiong, R., Ma, Z.-M., and Lan, Y.
\newblock When does group invariant learning survive spurious correlations?
\newblock \emph{Advances in Neural Information Processing Systems},
  35:\penalty0 7038--7051, 2022.

\bibitem[Dai \& Van~Gool(2018)Dai and Van~Gool]{dai2018dark}
Dai, D. and Van~Gool, L.
\newblock Dark model adaptation: Semantic image segmentation from daytime to
  nighttime.
\newblock In \emph{2018 21st International Conference on Intelligent
  Transportation Systems (ITSC)}, pp.\  3819--3824. IEEE, 2018.

\bibitem[Dong et~al.(2022)Dong, Muhammad, Zhou, Xie, Hu, Yang, Bae, and
  Li]{dong2022zood}
Dong, Q., Muhammad, A., Zhou, F., Xie, C., Hu, T., Yang, Y., Bae, S.-H., and
  Li, Z.
\newblock Zood: Exploiting model zoo for out-of-distribution generalization.
\newblock \emph{arXiv preprint arXiv:2210.09236}, 2022.

\bibitem[Ericsson et~al.(2021)Ericsson, Gouk, and
  Hospedales]{Ericsson2021HowTransfer}
Ericsson, L., Gouk, H., and Hospedales, T.~M.
\newblock How well do self-supervised models transfer?
\newblock In \emph{Proceedings of the IEEE/CVF Conference on Computer Vision
  and Pattern Recognition}, pp.\  5414--5423, 2021.

\bibitem[Fang et~al.(2013)Fang, Xu, and Rockmore]{Fang2013UnbiasedML}
Fang, C., Xu, Y., and Rockmore, D.~N.
\newblock Unbiased metric learning: On the utilization of multiple datasets and
  web images for softening bias.
\newblock \emph{2013 IEEE International Conference on Computer Vision}, pp.\
  1657--1664, 2013.

\bibitem[Gontijo-Lopes et~al.(2022)Gontijo-Lopes, Dauphin, and
  Cubuk]{gontijo-lopes2022no}
Gontijo-Lopes, R., Dauphin, Y., and Cubuk, E.~D.
\newblock No one representation to rule them all: Overlapping features of
  training methods.
\newblock In \emph{International Conference on Learning Representations}, 2022.
\newblock URL \url{https://openreview.net/forum?id=BK-4qbGgIE3}.

\bibitem[Gretton et~al.(2007)Gretton, Fukumizu, Teo, Song, Sch{\"o}lkopf, and
  Smola]{gretton2007kernel}
Gretton, A., Fukumizu, K., Teo, C., Song, L., Sch{\"o}lkopf, B., and Smola, A.
\newblock A kernel statistical test of independence.
\newblock \emph{Advances in neural information processing systems}, 20, 2007.

\bibitem[Grill et~al.(2020)Grill, Strub, Altch'e, Tallec, Richemond,
  Buchatskaya, Doersch, Pires, Guo, Azar, Piot, Kavukcuoglu, Munos, and
  Valko]{Grill2020BootstrapYO}
Grill, J.-B., Strub, F., Altch'e, F., Tallec, C., Richemond, P.~H.,
  Buchatskaya, E., Doersch, C., Pires, B.~{\'A}., Guo, Z.~D., Azar, M.~G.,
  Piot, B., Kavukcuoglu, K., Munos, R., and Valko, M.
\newblock Bootstrap your own latent: A new approach to self-supervised
  learning.
\newblock \emph{ArXiv}, abs/2006.07733, 2020.

\bibitem[Gulrajani \& Lopez-Paz(2021)Gulrajani and
  Lopez-Paz]{gulrajani2020search}
Gulrajani, I. and Lopez-Paz, D.
\newblock In search of lost domain generalization.
\newblock In \emph{International Conference on Learning Representations}, 2021.

\bibitem[He et~al.(2016)He, Zhang, Ren, and Sun]{He2016DeepRL}
He, K., Zhang, X., Ren, S., and Sun, J.
\newblock Deep residual learning for image recognition.
\newblock \emph{2016 IEEE Conference on Computer Vision and Pattern Recognition
  (CVPR)}, pp.\  770--778, 2016.

\bibitem[He et~al.(2021)He, Chen, Xie, Li, Doll{\'a}r, and
  Girshick]{he2021masked}
He, K., Chen, X., Xie, S., Li, Y., Doll{\'a}r, P., and Girshick, R.
\newblock Masked autoencoders are scalable vision learners.
\newblock \emph{arXiv preprint arXiv:2111.06377}, 2021.

\bibitem[Hu et~al.(2022{\natexlab{a}})Hu, Liu, Zhou, Wang, and
  Huang]{hu2022your}
Hu, T., Liu, Z., Zhou, F., Wang, W., and Huang, W.
\newblock Your contrastive learning is secretly doing stochastic neighbor
  embedding.
\newblock \emph{arXiv preprint arXiv:2205.14814}, 2022{\natexlab{a}}.

\bibitem[Hu et~al.(2022{\natexlab{b}})Hu, Wang, Wang, and
  Li]{hu2022understanding}
Hu, T., Wang, J., Wang, W., and Li, Z.
\newblock Understanding square loss in training overparametrized neural network
  classifiers.
\newblock \emph{Advances in Neural Information Processing Systems},
  35:\penalty0 16495--16508, 2022{\natexlab{b}}.

\bibitem[Huang et~al.(2017)Huang, Liu, and Weinberger]{Huang2017DenselyCC}
Huang, G., Liu, Z., and Weinberger, K.~Q.
\newblock Densely connected convolutional networks.
\newblock \emph{2017 IEEE Conference on Computer Vision and Pattern Recognition
  (CVPR)}, pp.\  2261--2269, 2017.

\bibitem[Idrissi et~al.(2022)Idrissi, Bouchacourt, Balestriero, Evtimov,
  Hazirbas, Ballas, Vincent, Drozdzal, Lopez-Paz, and
  Ibrahim]{idrissi2022imagenet}
Idrissi, B.~Y., Bouchacourt, D., Balestriero, R., Evtimov, I., Hazirbas, C.,
  Ballas, N., Vincent, P., Drozdzal, M., Lopez-Paz, D., and Ibrahim, M.
\newblock Imagenet-x: Understanding model mistakes with factor of variation
  annotations.
\newblock \emph{arXiv preprint arXiv:2211.01866}, 2022.

\bibitem[Inc.(2023)]{hughub}
Inc., H.~F.
\newblock The model hub of hugging face.
\newblock \url{https://huggingface.co/models}, 2023.

\bibitem[Kingma \& Ba(2014)Kingma and Ba]{kingma2014adam}
Kingma, D.~P. and Ba, J.
\newblock Adam: A method for stochastic optimization.
\newblock \emph{arXiv preprint arXiv:1412.6980}, 2014.

\bibitem[Krueger et~al.(2021)Krueger, Caballero, Jacobsen, Zhang, Binas, Zhang,
  Le~Priol, and Courville]{krueger2021out}
Krueger, D., Caballero, E., Jacobsen, J.-H., Zhang, A., Binas, J., Zhang, D.,
  Le~Priol, R., and Courville, A.
\newblock Out-of-distribution generalization via risk extrapolation (rex).
\newblock In \emph{International Conference on Machine Learning}, pp.\
  5815--5826. PMLR, 2021.

\bibitem[Kuang et~al.(2018)Kuang, Cui, Athey, Xiong, and Li]{Kuang2018}
Kuang, K., Cui, P., Athey, S., Xiong, R., and Li, B.
\newblock Stable prediction across unknown environments.
\newblock In \emph{Proceedings of the 24th ACM SIGKDD International Conference
  on Knowledge Discovery \& Data Mining}, pp.\  1617--1626, 2018.

\bibitem[Lee et~al.(2018)Lee, Lee, Lee, and Shin]{lee2018simple}
Lee, K., Lee, K., Lee, H., and Shin, J.
\newblock A simple unified framework for detecting out-of-distribution samples
  and adversarial attacks.
\newblock \emph{Advances in neural information processing systems}, 31, 2018.

\bibitem[Li et~al.(2017)Li, Yang, Song, and Hospedales]{Li2017DeeperBA}
Li, D., Yang, Y., Song, Y.-Z., and Hospedales, T.~M.
\newblock Deeper, broader and artier domain generalization.
\newblock \emph{2017 IEEE International Conference on Computer Vision (ICCV)},
  pp.\  5543--5551, 2017.

\bibitem[Li et~al.(2018)Li, Yang, Song, and Hospedales]{Li2018LearningTG}
Li, D., Yang, Y., Song, Y.-Z., and Hospedales, T.~M.
\newblock Learning to generalize: Meta-learning for domain generalization.
\newblock In \emph{Thirty-Second AAAI Conference on Artificial Intelligence},
  2018.

\bibitem[Li et~al.(2021)Li, Zhou, Xiong, Socher, and Hoi]{Li2021PrototypicalCL}
Li, J., Zhou, P., Xiong, C., Socher, R., and Hoi, S. C.~H.
\newblock Prototypical contrastive learning of unsupervised representations.
\newblock \emph{ArXiv}, abs/2005.04966, 2021.

\bibitem[Li et~al.(2022)Li, Ren, Jiang, Li, Zhang, and Li]{li2022domain}
Li, Z., Ren, K., Jiang, X., Li, B., Zhang, H., and Li, D.
\newblock Domain generalization using pretrained models without fine-tuning.
\newblock \emph{arXiv preprint arXiv:2203.04600}, 2022.

\bibitem[Liu et~al.(2021)Liu, Lin, Cao, Hu, Wei, Zhang, Lin, and
  Guo]{Liu2021SwinTH}
Liu, Z., Lin, Y., Cao, Y., Hu, H., Wei, Y., Zhang, Z., Lin, S., and Guo, B.
\newblock Swin transformer: Hierarchical vision transformer using shifted
  windows.
\newblock \emph{ArXiv}, abs/2103.14030, 2021.

\bibitem[Liu et~al.(2022)Liu, Han, Chen, Hong, Xu, Xu, and Li]{liu2022task}
Liu, Z., Han, J., Chen, K., Hong, L., Xu, H., Xu, C., and Li, Z.
\newblock Task-customized self-supervised pre-training with scalable dynamic
  routing.
\newblock \emph{Transfer}, 55:\penalty0 65, 2022.

\bibitem[Madry et~al.(2018)Madry, Makelov, Schmidt, Tsipras, and
  Vladu]{Madry2018TowardsDL}
Madry, A., Makelov, A., Schmidt, L., Tsipras, D., and Vladu, A.
\newblock Towards deep learning models resistant to adversarial attacks.
\newblock \emph{ArXiv}, abs/1706.06083, 2018.

\bibitem[Misra \& van~der Maaten(2020)Misra and van~der
  Maaten]{Misra2020SelfSupervisedLO}
Misra, I. and van~der Maaten, L.
\newblock Self-supervised learning of pretext-invariant representations.
\newblock \emph{2020 IEEE/CVF Conference on Computer Vision and Pattern
  Recognition (CVPR)}, pp.\  6706--6716, 2020.

\bibitem[Nagarajan et~al.(2021)Nagarajan, Andreassen, and
  Neyshabur]{nagarajan2021understanding}
Nagarajan, V., Andreassen, A., and Neyshabur, B.
\newblock Understanding the failure modes of out-of-distribution
  generalization.
\newblock In \emph{International Conference on Learning Representations}, 2021.

\bibitem[Paszke et~al.(2019)Paszke, Gross, Massa, Lerer, Bradbury, Chanan,
  Killeen, Lin, Gimelshein, Antiga, Desmaison, Kopf, Yang, DeVito, Raison,
  Tejani, Chilamkurthy, Steiner, Fang, Bai, and Chintala]{NEURIPS2019_9015}
Paszke, A., Gross, S., Massa, F., Lerer, A., Bradbury, J., Chanan, G., Killeen,
  T., Lin, Z., Gimelshein, N., Antiga, L., Desmaison, A., Kopf, A., Yang, E.,
  DeVito, Z., Raison, M., Tejani, A., Chilamkurthy, S., Steiner, B., Fang, L.,
  Bai, J., and Chintala, S.
\newblock Pytorch: An imperative style, high-performance deep learning library.
\newblock In Wallach, H., Larochelle, H., Beygelzimer, A., d`Alch\'{e} Buc, F.,
  Fox, E., and Garnett, R. (eds.), \emph{Advances in Neural Information
  Processing Systems 32}, pp.\  8024--8035. Curran Associates, Inc., 2019.

\bibitem[Peng et~al.(2019)Peng, Bai, Xia, Huang, Saenko, and
  Wang]{Peng2019MomentMF}
Peng, X., Bai, Q., Xia, X., Huang, Z., Saenko, K., and Wang, B.
\newblock Moment matching for multi-source domain adaptation.
\newblock \emph{2019 IEEE/CVF International Conference on Computer Vision
  (ICCV)}, pp.\  1406--1415, 2019.

\bibitem[Radford et~al.(2021)Radford, Kim, Hallacy, Ramesh, Goh, Agarwal,
  Sastry, Askell, Mishkin, Clark, Krueger, and
  Sutskever]{Radford2021LearningTV}
Radford, A., Kim, J.~W., Hallacy, C., Ramesh, A., Goh, G., Agarwal, S., Sastry,
  G., Askell, A., Mishkin, P., Clark, J., Krueger, G., and Sutskever, I.
\newblock Learning transferable visual models from natural language
  supervision.
\newblock In \emph{ICML}, 2021.

\bibitem[Ram{\'e} et~al.(2022)Ram{\'e}, Ahuja, Zhang, Cord, Bottou, and
  Lopez-Paz]{rame2022recycling}
Ram{\'e}, A., Ahuja, K., Zhang, J., Cord, M., Bottou, L., and Lopez-Paz, D.
\newblock Recycling diverse models for out-of-distribution generalization.
\newblock \emph{arXiv preprint arXiv:2212.10445}, 2022.

\bibitem[Rame et~al.(2022)Rame, Kirchmeyer, Rahier, Rakotomamonjy, patrick
  gallinari, and Cord]{rame2022diverse}
Rame, A., Kirchmeyer, M., Rahier, T., Rakotomamonjy, A., patrick gallinari, and
  Cord, M.
\newblock Diverse weight averaging for out-of-distribution generalization.
\newblock In Oh, A.~H., Agarwal, A., Belgrave, D., and Cho, K. (eds.),
  \emph{Advances in Neural Information Processing Systems}, 2022.
\newblock URL \url{https://openreview.net/forum?id=tq_J_MqB3UB}.

\bibitem[Russakovsky et~al.(2015{\natexlab{a}})Russakovsky, Deng, Su, Krause,
  Satheesh, Ma, Huang, Karpathy, Khosla, Bernstein,
  et~al.]{russakovsky2015imagenet}
Russakovsky, O., Deng, J., Su, H., Krause, J., Satheesh, S., Ma, S., Huang, Z.,
  Karpathy, A., Khosla, A., Bernstein, M., et~al.
\newblock Imagenet large scale visual recognition challenge.
\newblock \emph{International journal of computer vision}, 115\penalty0
  (3):\penalty0 211--252, 2015{\natexlab{a}}.

\bibitem[Russakovsky et~al.(2015{\natexlab{b}})Russakovsky, Deng, Su, Krause,
  Satheesh, Ma, Huang, Karpathy, Khosla, Bernstein, Berg, and
  Fei-Fei]{Russakovsky2015ImageNetLS}
Russakovsky, O., Deng, J., Su, H., Krause, J., Satheesh, S., Ma, S., Huang, Z.,
  Karpathy, A., Khosla, A., Bernstein, M.~S., Berg, A.~C., and Fei-Fei, L.
\newblock Imagenet large scale visual recognition challenge.
\newblock \emph{International Journal of Computer Vision}, 115:\penalty0
  211--252, 2015{\natexlab{b}}.

\bibitem[Salman et~al.(2020)Salman, Ilyas, Engstrom, Kapoor, and
  Madry]{Salman2020DoAR}
Salman, H., Ilyas, A., Engstrom, L., Kapoor, A., and Madry, A.
\newblock Do adversarially robust imagenet models transfer better?
\newblock \emph{ArXiv}, abs/2007.08489, 2020.

\bibitem[Sandler et~al.(2018)Sandler, Howard, Zhu, Zhmoginov, and
  Chen]{Sandler2018MobileNetV2IR}
Sandler, M., Howard, A.~G., Zhu, M., Zhmoginov, A., and Chen, L.-C.
\newblock Mobilenetv2: Inverted residuals and linear bottlenecks.
\newblock \emph{2018 IEEE/CVF Conference on Computer Vision and Pattern
  Recognition}, pp.\  4510--4520, 2018.

\bibitem[Szegedy et~al.(2015)Szegedy, Liu, Jia, Sermanet, Reed, Anguelov,
  Erhan, Vanhoucke, and Rabinovich]{Szegedy2015GoingDW}
Szegedy, C., Liu, W., Jia, Y., Sermanet, P., Reed, S.~E., Anguelov, D., Erhan,
  D., Vanhoucke, V., and Rabinovich, A.
\newblock Going deeper with convolutions.
\newblock \emph{2015 IEEE Conference on Computer Vision and Pattern Recognition
  (CVPR)}, pp.\  1--9, 2015.

\bibitem[Szegedy et~al.(2016)Szegedy, Vanhoucke, Ioffe, Shlens, and
  Wojna]{Szegedy2016RethinkingTI}
Szegedy, C., Vanhoucke, V., Ioffe, S., Shlens, J., and Wojna, Z.
\newblock Rethinking the inception architecture for computer vision.
\newblock \emph{2016 IEEE Conference on Computer Vision and Pattern Recognition
  (CVPR)}, pp.\  2818--2826, 2016.

\bibitem[Tan \& Le(2019)Tan and Le]{Tan2019EfficientNetRM}
Tan, M. and Le, Q.
\newblock Efficientnet: Rethinking model scaling for convolutional neural
  networks.
\newblock In \emph{International Conference on Machine Learning}, pp.\
  6105--6114. PMLR, 2019.

\bibitem[Venkateswara et~al.(2017)Venkateswara, Eusebio, Chakraborty, and
  Panchanathan]{Venkateswara2017DeepHN}
Venkateswara, H., Eusebio, J., Chakraborty, S., and Panchanathan, S.
\newblock Deep hashing network for unsupervised domain adaptation.
\newblock \emph{2017 IEEE Conference on Computer Vision and Pattern Recognition
  (CVPR)}, pp.\  5385--5394, 2017.

\bibitem[Volk et~al.(2019)Volk, M{\"u}ller, Von~Bernuth, Hospach, and
  Bringmann]{volk2019towards}
Volk, G., M{\"u}ller, S., Von~Bernuth, A., Hospach, D., and Bringmann, O.
\newblock Towards robust cnn-based object detection through augmentation with
  synthetic rain variations.
\newblock In \emph{2019 IEEE Intelligent Transportation Systems Conference
  (ITSC)}, pp.\  285--292. IEEE, 2019.

\bibitem[Wang et~al.(2022)Wang, Yi, Chen, and Zhu]{wang2022out}
Wang, R., Yi, M., Chen, Z., and Zhu, S.
\newblock Out-of-distribution generalization with causal invariant
  transformations.
\newblock In \emph{Proceedings of the IEEE/CVF Conference on Computer Vision
  and Pattern Recognition}, pp.\  375--385, 2022.

\bibitem[Wiles et~al.(2022)Wiles, Gowal, Stimberg, Rebuffi, Ktena, Dvijotham,
  and Cemgil]{wiles2022a}
Wiles, O., Gowal, S., Stimberg, F., Rebuffi, S.-A., Ktena, I., Dvijotham,
  K.~D., and Cemgil, A.~T.
\newblock A fine-grained analysis on distribution shift.
\newblock In \emph{International Conference on Learning Representations}, 2022.
\newblock URL \url{https://openreview.net/forum?id=Dl4LetuLdyK}.

\bibitem[Wolf et~al.(2020)Wolf, Debut, Sanh, Chaumond, Delangue, Moi, Cistac,
  Rault, Louf, Funtowicz, Davison, Shleifer, von Platen, Ma, Jernite, Plu, Xu,
  Scao, Gugger, Drame, Lhoest, and Rush]{wolf-etal-2020-transformers}
Wolf, T., Debut, L., Sanh, V., Chaumond, J., Delangue, C., Moi, A., Cistac, P.,
  Rault, T., Louf, R., Funtowicz, M., Davison, J., Shleifer, S., von Platen,
  P., Ma, C., Jernite, Y., Plu, J., Xu, C., Scao, T.~L., Gugger, S., Drame, M.,
  Lhoest, Q., and Rush, A.~M.
\newblock Transformers: State-of-the-art natural language processing.
\newblock In \emph{Proceedings of the 2020 Conference on Empirical Methods in
  Natural Language Processing: System Demonstrations}, pp.\  38--45, Online,
  October 2020. Association for Computational Linguistics.
\newblock URL \url{https://www.aclweb.org/anthology/2020.emnlp-demos.6}.

\bibitem[Wortsman et~al.(2022)Wortsman, Ilharco, Gadre, Roelofs, Gontijo-Lopes,
  Morcos, Namkoong, Farhadi, Carmon, Kornblith, et~al.]{wortsman2022model}
Wortsman, M., Ilharco, G., Gadre, S.~Y., Roelofs, R., Gontijo-Lopes, R.,
  Morcos, A.~S., Namkoong, H., Farhadi, A., Carmon, Y., Kornblith, S., et~al.
\newblock Model soups: averaging weights of multiple fine-tuned models improves
  accuracy without increasing inference time.
\newblock In \emph{International Conference on Machine Learning}, pp.\
  23965--23998. PMLR, 2022.

\bibitem[Wu et~al.(2020)Wu, Xu, Dai, Wan, Zhang, Yan, Tomizuka, Gonzalez,
  Keutzer, and Vajda]{wu2020visual}
Wu, B., Xu, C., Dai, X., Wan, A., Zhang, P., Yan, Z., Tomizuka, M., Gonzalez,
  J., Keutzer, K., and Vajda, P.
\newblock Visual transformers: Token-based image representation and processing
  for computer vision, 2020.

\bibitem[Wu et~al.(2018)Wu, Xiong, Yu, and Lin]{Wu2018UnsupervisedFL}
Wu, Z., Xiong, Y., Yu, S.~X., and Lin, D.
\newblock Unsupervised feature learning via non-parametric instance
  discrimination.
\newblock In \emph{Proceedings of the IEEE conference on computer vision and
  pattern recognition}, pp.\  3733--3742, 2018.

\bibitem[Xie et~al.(2017)Xie, Girshick, Doll{\'a}r, Tu, and
  He]{Xie2017AggregatedRT}
Xie, S., Girshick, R.~B., Doll{\'a}r, P., Tu, Z., and He, K.
\newblock Aggregated residual transformations for deep neural networks.
\newblock \emph{2017 IEEE Conference on Computer Vision and Pattern Recognition
  (CVPR)}, pp.\  5987--5995, 2017.

\bibitem[Xiong et~al.(2021)Xiong, Chen, Pang, Cheng, Ma, and
  Lan]{xiong2021uncertainty}
Xiong, R., Chen, Y., Pang, L., Cheng, X., Ma, Z.-M., and Lan, Y.
\newblock Uncertainty calibration for ensemble-based debiasing methods.
\newblock \emph{Advances in Neural Information Processing Systems},
  34:\penalty0 13657--13669, 2021.

\bibitem[Yalniz et~al.(2019)Yalniz, J{\'e}gou, Chen, Paluri, and
  Mahajan]{Yalniz2019BillionscaleSL}
Yalniz, I.~Z., J{\'e}gou, H., Chen, K., Paluri, M., and Mahajan, D.~K.
\newblock Billion-scale semi-supervised learning for image classification.
\newblock \emph{ArXiv}, abs/1905.00546, 2019.

\bibitem[Ye et~al.(2022)Ye, Li, Bai, Yu, Hong, Zhou, Li, and Zhu]{ye2022ood}
Ye, N., Li, K., Bai, H., Yu, R., Hong, L., Zhou, F., Li, Z., and Zhu, J.
\newblock Ood-bench: Quantifying and understanding two dimensions of
  out-of-distribution generalization.
\newblock In \emph{Proceedings of the IEEE/CVF Conference on Computer Vision
  and Pattern Recognition}, pp.\  7947--7958, 2022.

\bibitem[Yi et~al.(2021)Yi, Hou, Sun, Shang, Jiang, Liu, and
  Ma]{yi2021improved}
Yi, M., Hou, L., Sun, J., Shang, L., Jiang, X., Liu, Q., and Ma, Z.
\newblock Improved ood generalization via adversarial training and pretraing.
\newblock In \emph{International Conference on Machine Learning}, pp.\
  11987--11997. PMLR, 2021.

\bibitem[Yi et~al.(2023)Yi, Wang, Sun, Li, and Ma]{yi2023breaking}
Yi, M., Wang, R., Sun, J., Li, Z., and Ma, Z.-M.
\newblock Breaking correlation shift via conditional invariant regularizer.
\newblock In \emph{The Eleventh International Conference on Learning
  Representations}, 2023.

\bibitem[You et~al.(2021)You, Liu, Wang, and Long]{you2021logme}
You, K., Liu, Y., Wang, J., and Long, M.
\newblock Logme: Practical assessment of pre-trained models for transfer
  learning.
\newblock In \emph{International Conference on Machine Learning}, pp.\
  12133--12143. PMLR, 2021.

\end{thebibliography}
\bibliographystyle{icml2023}

\clearpage
\appendix
\onecolumn

\section{Shift metrics}

\subsection{Practical Estimation}
\label{apsubsec:estimation}

In this section, we show how the two metrics can be computed practically for general latent features of an arbitrary PTM. 

\paragraph{Notations.} Consider a dataset $\mathcal{D}$ that contains samples collected under multiple domains $\mathcal{E}$, i.e., $\mathcal{D} = \{D_e\}_{e\in \mathcal{E}}$. $D_e=\left\{x_{i}^{e}, y_{i}^{e}\right\}_{i=1}^{n^{e}}$ contains instances of random variables $(X, Y)$ that are \emph{i.i.d.} sampled from the probability distribution $\mathbb{P}^e(\mathcal{X} \times \mathcal{Y})$. A PTM is denoted as a feature encoder $\phi: \mathcal{X} \rightarrow \mathcal{Z}_{\phi}$. Suppose the dimension of $\phi(x)$ is $d$. The feature matrix on the domain $D_e$ is denoted as
\[ \Phi_e := \big( \phi(x^e_{1}), \phi(x^e_{i2}), \dots, \phi(x^e_{n^e})  \big)^\top \in \mathbb{R}^{n^e \times d}. \]

\paragraph{Diversity shift.} 
Denote $\mathcal{S}_{e}(e', \phi) := \{\mathbf{z} \in \mathcal{Z}_{\phi}| p_e(\mathbf{z}) > 0, p_{e'}(\mathbf{z}) = 0 \}$, $\mathcal{S}_{e'}(e, \phi) := \{\mathbf{z} \in \mathcal{Z}_{\phi}| p_{e}(\mathbf{z}) = 0, p_{e'}(\mathbf{z}) > 0 \}$,
$F_{div}(\phi, e, e')$ can be written as
\begin{align*}
 F_{div}(\phi, e, e') = 
    \frac{1}{2} ( \mathbb{P}^e[\mathcal{S}_{e}(e', \phi)] + \mathbb{P}^{e'}[\mathcal{S}_{e'}(e, \phi)]).
\end{align*}
We design the following empirical estimation of $\mathbb{P}^e[\mathcal{S}_{e}(e', \phi)]$:
\begin{align*}
    \hat{\mathbb{P}}^{e}[\hat{\mathcal{S}_{e}}(e', \phi)] := \hat{\mathbb{P}}^{e} (\{ \mathbf{x} \in D_e | \hat{p}_{e'}(\mathbf{z}) < \epsilon_{e'}, \mathbf{z} = \phi(\mathbf{x}) \}).
\end{align*}
Intuitively, we estimate the no-overlap set $\mathcal{S}_e(e', \phi)$ using the estimated probability of the instance in the estimated distribution $\hat{p}_{e'}$. When the probability is lower than a given small threshold $\epsilon_{e'}$, it is considered as in the set $\mathcal{S}_e(e', \phi)$. The threshold $\epsilon_{e'}$ is estimated by
\begin{align*}
    \hat{\mathbb{P}}^{e'} (\{ \mathbf{x} \in V_{e'} | \hat{p}_{e'}(\mathbf{z}) < \epsilon_{e'}, \mathbf{z} = \phi(\mathbf{x}) \}) = 0.01.
\end{align*}
For each $e \in \mathcal{E}$, we approximate $p_{e}$ with a Gaussian distribution $\mathcal{N}(\mu_{e}, \Sigma_e)$, and estimate the parameters with empirical statistics on $D_e$. Specifically,
\begin{align*}
    \hat \mu_e = \frac{1}{n^e} \Phi_{e}^\top \mathbbm{1}_{n^e} \quad \hat \Sigma_e = \frac{1}{n^e} (\Phi_{e} - \mathbbm{1}_{n^e} \hat \mu_\phi^\top )^\top (\Phi_{e} - \mathbbm{1}_{n^e} \hat \mu_\phi^\top ),
\end{align*}

Given the estimated distribution $\mathcal{N}(\hat \mu_{e}, \hat \Sigma_e)$, the probability density at a given point $\mathbf{z} \in \mathcal{Z}_\phi$ is computed as
\begin{align*}
    \hat p_e(\mathbf{z}) = \hat p_e(\mathbf{z}|\hat \mu_e, \hat \Sigma_e) = \sqrt{\frac{1}{(2\pi)^d |\hat \Sigma_e| }} \exp\left(-\frac{1}{2}(\mathbf{z}-\hat \mu_e)^\top \hat \Sigma_e^{-1} (\mathbf{z}-\hat \mu_e) \right).
\end{align*}
Denote $C_e := (2\pi)^{-d/2} |\hat \Sigma_e|^{1/2}$, $\hat d_e(\mathbf{z}) := (\mathbf{z}-\hat \mu_e)^\top \hat \Sigma_e^{-1} (\mathbf{z}-\hat \mu_e)$, we have $\hat p_e (\mathbf{z}) = C_e \exp(-\frac{1}{2} \hat d_e(\mathbf{z}))$.
As $C_e$ is constant for any $\mathbf{z}$, and the exponential function is monotonic, we can empirically estimate $\mathbb{P}^e[\mathcal{S}_{e}(e', \phi)]$ using $\hat d_{e'}$ instead as follows:
\begin{align*}
    \hat{\mathbb{P}}^{e}[\hat{\mathcal{S}_{e}}(e', \phi)] := \hat{\mathbb{P}}^{e} (\{ \mathbf{x} \in D_e | \hat{d}_{e'}(\mathbf{z}) > \epsilon_{e'}, \mathbf{z} = \phi(\mathbf{x}) \}),
\end{align*}
where $\epsilon_{e'}$ satisfies
\begin{equation*}
    \hat{\mathbb{P}}^{e'} (\{ \mathbf{x} \in V_{e'} | \hat{d}_{e'}(\mathbf{z}) > \epsilon_{e'}, \mathbf{z} = \phi(\mathbf{x}) \}) = 0.01.
\end{equation*}
Note that it connects to the common practice in OOD detection methods where the Mahalanobis distance is estimated~\cite{lee2018simple}. \\

The estimation of $\mathbb{P}^{e'}[\mathcal{S}_{e'}(e, \phi)]$ is defined in the same way. The empirical diversity metric is then the average of the two estimations, i.e., 
\begin{equation*}
    \hat F_{div}(\phi, e, e') = 
    \frac{1}{2} \left( \hat{\mathbb{P}}^e[\hat{\mathcal{S}}_{e}(e', \phi)] + \hat{\mathbb{P}}^{e'}[\hat{\mathcal{S}}_{e'}(e, \phi)] \right).
\end{equation*}

\paragraph{Correlation shift.} For each pair of domain $e, e'$. We have the empirical set $\hat{\mathcal{T}}(\phi, e, e') := (D_e \setminus \hat{S}_e(e', \phi)) \cup (D_{e'} \setminus \hat{S}_{e'}(e, \phi)) $. Denote $p_{e, e'} = \frac{1}{2}(p_e + p_{e'})$ and 
\begin{align*}
    \hat{D}_{cor} &= \frac{1}{2} \sum_{\mathbf{x} \in \hat{\mathcal{T}}} \hat{p}_{e, e'}(\mathbf{x}) \sum_{y \in \mathcal{Y}} |\hat{p}_{e}(y|\phi(\mathbf{x})) - \hat{p}_{e'}(y|\phi(\mathbf{x}))|.
\end{align*}
As $D_e, D_{e'}$ are independently sampled, $\hat{p}_{e, e'}(\mathbf{x})$ can be estimated by the empirical distribution, i.e., $\hat{p}_{e, e'}(\mathbf{x}) = 1 / |D_e \cup D_e'| $. To estimate $\hat{p}_{e}(y|\phi(\mathbf{x}))$,
we first get a primary estimation $\tilde{p}_{e}(y|\phi(\mathbf{x}))$ with the following equation:
\begin{align}
\label{eq:estimate_p_e}
    \tilde{p}_{e}(\mathbf{y}|\phi(\mathbf{x})) := m(\mathbf{M}_0\phi(\mathbf{x}), \mathbf{M}_1\phi(\mathbf{x}),\dots,\mathbf{M}_{|\mathcal{Y}|}\phi(\mathbf{x})),
\end{align}
where $m$ denotes the normalization operator. The coefficient matrices $(\mathbf{M}_0, \mathbf{M}_1, \dots,\mathbf{M}_{|\mathcal{Y}|})$ are estimated by minimizing the empirical evidence on $D_e$ as in LogME~\cite{you2021logme}. Specifically, denote $K:=|\mathcal{Y}|$, $\mathbf{y} \in \mathbb{R}^{K}$ is the one-hot label vector . Denote $y_i$ as the $i$-th component. We adopt the following linear model assumption:
\begin{equation*}
    y_i = \mathbf{w}_i^\top \phi(x) + \epsilon, \quad \mathbf{w}_i \in \mathbb{R}^{d}, \,\,  \epsilon \in \mathbb{R},
\end{equation*}
where $\epsilon$ is the Gaussian noise variable with variance $\beta^{-1}$. As we assume that the prior distribution of weights $\mathbf{w}_i$ is an isotropic Gaussian distribution with zero mean and parameterized by $\alpha$, i.e.
\begin{align*}
    \mathbf{w}_i \sim \mathcal{N}(\mathbf{0}, \alpha^{-1} \mathbb{I}_d),
\end{align*}
and the conditional distribution of 
$y_i$ given $\phi(x)$ is
\begin{align*}
    y_i \big|\phi(x), \mathbf{w}_i \sim \mathcal{N}(\mathbf{w}_i^\top \phi(x), \beta^{-1}),
\end{align*}
then according to the definition of evidence,
\begin{align*}
    p(y_i | \phi(x), \alpha, \beta) = \int_{\mathbf{w}_i \in \mathbb{R}^{d}} p(\mathbf{w}_i|\alpha) p(y_i | \phi(x), \mathbf{w}_i, \beta) d\mathbf{w}_i.
\end{align*}
Denote $\Phi_e \in \mathbb{R}^{n_e \times d}$ as the feature matrix of all training samples in the environment $e$, and $\mathbf{y}_i^e \in \mathbb{R}^{n_e}$ as the label vector composed by $y_i$. Denote $A = \alpha I + \beta \Phi_e^{\top}\Phi_e, m = \beta A^{-1}\Phi_e^{\top}\mathbf{y}_i^e$, we have the following log-likelihood:
\begin{align*}
    \mathcal{L}(\alpha, \beta) & =\log p(\mathbf{y}_i^e | \Phi_e, \alpha, \beta) \\
    & =\frac{n_e}{2} \log \beta+\frac{d}{2} \log \alpha-\frac{n_e}{2} \log 2 \pi \\ & -\frac{\beta}{2}\| \Phi_e m-\mathbf{y}_i^e\|_2^2-\frac{\alpha}{2} m^{\top} m-\frac{1}{2} \log |A|.
\end{align*}
Solve $(\alpha^*, \beta^*) = \arg \max_{\alpha, \beta} \mathcal{L}(\alpha, \beta)$ by using the same iterative approach as in ~\citep{you2021logme}, we can get an estimate of $\mathbf{w}_i$:
\begin{equation*}
    \hat{\mathbf{w}}_i = \beta^* (\alpha^* I + \beta^* \Phi_e^{\top}\Phi_e)^{-1}\Phi_e^{\top}\mathbf{y}_i^e.
\end{equation*}
Substituting the above estimate into the formula~\ref{eq:estimate_p_e}, we get the estimate $\tilde{p}_{e}(y|\phi(\mathbf{x}))$.
Alternatively, we can also consider directly using square loss for classifying the features and estimating the conditional probability \citep{hu2022understanding}. 
We then calibrate $\tilde{p}_{e}(y|\phi(\mathbf{x}))$ with the empirical accuracy estimated on $\hat{\mathcal{T}}(\phi, e, e')$ to get the final estimation $\hat{p}_{e}(y|\phi(\mathbf{x}))$. Specifically, denote $\mathcal{B}_0, \mathcal{B}_1, \dots, \mathcal{B}_b$ as $b$ average sized blocks in $[0, 1]$. We define
\begin{equation*}
    \mathcal{B}_i(\phi, e, y) = \{(\mathbf{x},y) \in D_e \cap \hat{T}(\phi, e, e') | \tilde{p}_e(y|\phi(\mathbf{x})) \in \mathcal{B}_i \}.
\end{equation*}
We then estimate $\hat{p}_e(y|\phi(\mathbf{x}))$ for $\mathbf{x} \in \mathcal{B}_i(\phi, e, y)$ as follows:
\begin{align*}
    \hat{p}_e(y|\phi(\mathbf{x})) := \frac{|\{(\mathbf{x}, y_x) \in \mathcal{B}_i(\phi, e, y)| y_x = y \}|}{|\mathcal{B}_i(\phi, e, y)|}.
\end{align*}
The final $\hat{D}_{cor}$ is then computed with the estimated $\hat{p}_{e}(y|\phi(\mathbf{x}))$ and $\hat{p}_{e'}(y|\phi(\mathbf{x}))$.

\begin{table*}[t]
    \centering
    \caption{Details of the model zoo proposed in~\cite{dong2022zood}. The first column corresponds to the numbers we have used for subsequent tables. The rest of the table describes architectures, pre-training datasets, and pre-training algorithms as well as the group and source of each model.}
    \label{tab_supp:pre-trained_models}
    \begin{adjustbox}{width=1.0\textwidth}
        \begin{tabular}{clllll}
        \toprule
        Number & Architecture & Dataset & Algorithm & Group & Source  \\
        \hline
        1 & ResNet-50 & ImageNet-1K    & ERM & Group 1 & \citet{NEURIPS2019_9015}\\
        2 & ResNet-152 & ImageNet-1K   & ERM & Group 1 & \citet{NEURIPS2019_9015}\\
        3 & ResNeXt-50 & ImageNet-1K   & ERM & Group 1 & \citet{NEURIPS2019_9015}\\
        4 & DenseNet-169 & ImageNet-1K & ERM & Group 1 & \citet{NEURIPS2019_9015}\\
        5 & DenseNet-201 & ImageNet-1K & ERM & Group 1 & \citet{NEURIPS2019_9015}\\
        6 & Inception v1 & ImageNet-1K & ERM & Group 1 & \citet{NEURIPS2019_9015} \\
        7 & Inception v3 & ImageNet-1K & ERM & Group 1 & \citet{NEURIPS2019_9015}\\
        8 & MobileNet v2 & ImageNet-1K & ERM & Group 1 & \citet{NEURIPS2019_9015}\\
        9 & EfficientNet-B2 & ImageNet-1K & ERM & Group 1 & \citet{NEURIPS2019_9015}\\
        10 & EfficientNet-B4 & ImageNet-1K & ERM & Group 1 & \citet{NEURIPS2019_9015} \\
        11 & Swin-T & ImageNet-1K & Swin & Group 1 & \citet{Liu2021SwinTH}\\
        12 & Swin-B & ImageNet-1K & Swin & Group 1 & \citet{Liu2021SwinTH}\\
        \hline
        13 & ResNet-50 & ImageNet-1K & Adv. $\ell_2$ ($\epsilon=0.5$) & Group 2& \citet{Salman2020DoAR} \\
        14 & ResNet-50 & ImageNet-1K & Adv. $\ell_\infty$ ($\epsilon=4$) & Group 2& \citet{Salman2020DoAR} \\
        15 & ResNet-50 & ImageNet-1K & BYOL & Group 2 & \citet{Ericsson2021HowTransfer}\\
        16 & ResNet-50 & ImageNet-1K & MoCo-v2 & Group 2 & \citet{Ericsson2021HowTransfer}\\
        17 & ResNet-50 & ImageNet-1K & InsDis & Group 2 & \citet{Ericsson2021HowTransfer}\\
        18 & ResNet-50 & ImageNet-1K & PIRL & Group 2 & \citet{Ericsson2021HowTransfer}\\
        19 & ResNet-50 & ImageNet-1K & DeepCluster-v2 & Group 2 & \citet{Ericsson2021HowTransfer}\\
        20 & ResNet-50 & ImageNet-1K & PCL-v2 & Group 2 &  \citet{Ericsson2021HowTransfer}\\
        21 & ResNet-50 & ImageNet-1K & SeLa-v2 & Group 2 & \citet{Ericsson2021HowTransfer}\\
        22 & ResNet-50 & ImageNet-1K & SwAV & Group 2 & \citet{Ericsson2021HowTransfer}\\
        \hline
        23 & ResNet-18   & ImageNet-1K + YFCC-100M  & Semi-supervised & Group 3 & \citet{Yalniz2019BillionscaleSL} \\
        24 & ResNet-50   & ImageNet-1K + YFCC-100M  & Semi-supervised & Group 3 & \citet{Yalniz2019BillionscaleSL} \\
        25 & ResNeXt-50  & ImageNet-1K + YFCC-100M  & Semi-supervised & Group 3 & \citet{Yalniz2019BillionscaleSL} \\
        26 & ResNeXt-101 & ImageNet-1K + YFCC-100M  & Semi-supervised & Group 3 & \citet{Yalniz2019BillionscaleSL} \\
        \hline
        27 & ResNet-18   & ImageNet-1K + IG-1B-Targeted & Semi-weakly Supervised & Group 3    & \citet{Yalniz2019BillionscaleSL}\\
        28 & ResNet-50   & ImageNet-1K + IG-1B-Targeted & Semi-weakly Supervised & Group 3    & \citet{Yalniz2019BillionscaleSL}\\
        29 & ResNeXt-50  & ImageNet-1K + IG-1B-Targeted & Semi-weakly Supervised & Group 3    & \citet{Yalniz2019BillionscaleSL}\\
        30 & ResNeXt-101 & ImageNet-1K + IG-1B-Targeted & Semi-weakly Supervised & Group 3    & \citet{Yalniz2019BillionscaleSL}\\
        \hline
        31 & Swin-B & ImageNet-1K + ImageNet-22K & Swin & Group 3 & \citet{Liu2021SwinTH}\\
        32 & BEiT-B   & ImageNet-1K + ImageNet-22K & BEiT & Group 3 & \citet{wolf-etal-2020-transformers,bao2021beit}\\
        33 & ViT-B/16 & ImageNet-1K + ImageNet-22K & ViT & Group 3 & \citet{wolf-etal-2020-transformers,wu2020visual}\\
        \hline
        34 & ResNet-50 & WebImageText & CLIP & Group 3 & \citet{Radford2021LearningTV}\\
        35 & ViT-B/16 & WebImageText & CLIP & Group 3 & \citet{Radford2021LearningTV}\\
        \bottomrule
        \end{tabular}
    \end{adjustbox}
\end{table*}

\begin{table*}
    \centering
    \caption{Dataset Statistics and model information. \textit{feature dim.} denotes the feature dimension of each model. Results denoted with ${\dagger}$ are from~\cite{dong2022zood}.}
    \label{aptab:datasets}
    \begin{tabular}{l|ccccc}
        \toprule
        \textbf{Datasets} &  \textbf{PACS}  & \textbf{VLCS}  & \textbf{OfficeHome} & \textbf{TerraInc.} & \textbf{DomainNet} \\
        \midrule
        \#samples & 7995 & 8584 & 12472 & 19832 & 469262 \\
        \#domains & 4 & 4 & 4 & 4 & 6 \\
        \#classes & 7 & 5 & 65 & 10 & 345 \\
        \midrule
        Main model & CLIP-ViT & CLIP-ViT & Swin-B-22 & Swin-B-22 & ResNext-101 \\
        feature dim. & 512 & 512 & 1024 & 1024 & 2048 \\
        finetuned$^{\dagger}$ & 96.0 & 79.5 & 84.6 & 37.3 & 48.8 \\
        rank (ZooD)$^{\dagger}$ & 1 & 1 & 1 & 1 & 1 \\
        \midrule
        HSIC aux. & ResNet50-ss & ResNet50-InsDis & ResNet50-InsDis & ResNet50-PIRL & ViT-B \\
        feature dim. & 2048 & 2048 & 2048 & 2048 & 768 \\
        finetuned$^{\dagger}$ & 75.7 & 65.6 & 22.7 & 18.4 & 34.1 \\
        rank (ZooD)$^{\dagger}$ & 6	& 30 & 34 & 32 & 13 \\
        \midrule
        Rew. aux. & BEiT-base & BEiT-base & deepcluster-v2 & inception-v3 & ResNet50-sws \\
        feature dim. & 768 & 768 & 2048 & 2048 & 2048 \\
        finetuned$^{\dagger}$ & 47.1 & 68.4 & 61.0 & 23.8 & 46.3 \\
        rank (ZooD)$^{\dagger}$ & 31 & 34 & 24 & 25 & 3 \\
        \bottomrule
    \end{tabular}
\end{table*}

\subsection{The Model Zoo}
\label{apsubsec:models}

We follow the model zoo setting of ZooD~\cite{dong2022zood} which consists of 35 PTMs having diverse architectures, pre-training methods, and pre-training datasets. A summary of the PTMs can be found in Table~\ref{tab_supp:pre-trained_models}. Dong et al.~(\citeyear{dong2022zood}) divide the models into three groups. In the main paper, we also introduce 3 subsets of models, with results shown in Figure~\ref{fig:ermcnn},~\ref{fig:rn50}, and~\cref{fig:swin}, respectively. \par

\textbf{Figure~\ref{fig:ermcnn}} contains results for models of 10 different architectures (CNNs) trained on ImageNet-1K with ERM. The architectures are as follows: ResNet-50, ResNet-152~\cite{He2016DeepRL}, 
ResNeXt-50~\cite{Xie2017AggregatedRT}, DenseNet-169, DenseNet-201~\cite{Huang2017DenselyCC}, Inception v1~\cite{Szegedy2015GoingDW}, Inception v3~\cite{Szegedy2016RethinkingTI}, MobileNet v2~\cite{Sandler2018MobileNetV2IR}, EfficientNet-B2, EfficientNet-B4~\cite{Tan2019EfficientNetRM}. \par

\textbf{Figure~\ref{fig:rn50}} contains 10 ResNet-50s trained via following pre-training methods: Adversarial Training~\cite{Madry2018TowardsDL}, BYOL~\cite{Grill2020BootstrapYO}, MoCo-v2~\cite{Chen2020ImprovedBW}, InsDis~\cite{Wu2018UnsupervisedFL}, PIRL~\cite{Misra2020SelfSupervisedLO}, DeepCluster-v2~\cite{Caron2018DeepCF}, PCL-v2~\cite{Li2021PrototypicalCL}, SeLa-v2~\cite{Asano2020SelflabellingVS, Caron2020UnsupervisedLO}, SwAV~\cite{Caron2020UnsupervisedLO}. \par

\textbf{Figure~\ref{fig:swin}} shows the results of 2 different versions of Swin-B~\cite{Liu2021SwinTH} pre-trained on ImageNet-1K or on both ImageNet-1K and ImageNet-22K~\cite{Russakovsky2015ImageNetLS}.

\section{Experiments}

\subsection{Experiment Details}
\label{apsubsec:config}

\textbf{Datasets.} Details of the five datasets in our experiments are introduced as follows.
\textbf{PACS}~\citep{Li2017DeeperBA}: This dataset contains a total of 9,991 images, drawn from four distinct domains (art, cartoons, photos, sketches), and encapsulates seven different classes.
\textbf{VLCS}~\citep{Fang2013UnbiasedML}: This compilation features 10,729 images from four domains (Caltech101, LabelMe, SUN09, VOC2007), comprising five distinct classes.
\textbf{Office-Home}~\citep{Venkateswara2017DeepHN}: This dataset includes images from four domains (art, clipart, product, real), primarily illustrating common objects in office and home environments. It is composed of a total of 15,588 images distributed across 65 classes.
\textbf{TerraIncognita}~\citep{Beery2018RecognitionIT}: This dataset encompasses photographs of wildlife captured by camera traps at four different locations. It contains a total of 24,788 images across 10 classes.
\textbf{DomainNet}~\citep{Peng2019MomentMF}: Recognized as one of the most challenging DG datasets, it comprises 586,575 images from six diverse domains (clipart, infographics, painting, quickdraw, real, sketch), spanning 345 classes.

\paragraph{Main and auxiliary models.} For the main model, we use the Top-1 model ranked by ZooD. For the auxiliary model, we select models that are extreme at one shift metric on that dataset. Table~\ref{aptab:datasets} shows some detailed statistics of selected auxiliary models and the main models. \textit{finetuned} denotes the averaged OOD accuracy of the linear classifier on the top of the corresponding PTM when finetuned on the dataset. \textit{rank (ZooD)} denotes the rank of the PTM according to the ZooD evaluation metric. According to the results in Table~\ref{aptab:datasets}, the finetuned performance of the selected auxiliary models are mostly weak.

We use a 3 layers MLP as the prediction head on the top of the main model and fine-tune it on the downstream tasks. The dimension of the first hidden layer is half of that of the output of the main model. The dimension of the second layer is set to 256 for all the main models except for ResNext-101, which is set to 512. The last layer is linear with the outputs of the same dimension as the class numbers. For the reweight auxiliary model, we use a linear layer on top of it and fine-tune it as the classifier. The reweight auxiliary classifier is trained under the following hyperparameter setting: $\textrm{learning rate}=1\times 10^{-5}, \textrm{batch size}=16, \textrm{dropout}=0, \textrm{weight decay}=0, \textrm{steps}=1000$. For DomainNet, the training steps are increased to $5000$.

\clearpage

\begin{table*}[t]
    \caption{Hyperparameters, their default values, and the search range.} 
    \begin{center}
    { 
    \begin{tabular}{llll}
        \toprule
        \textbf{Dataset} & \textbf{Parameter} & \textbf{Default value} & \textbf{Range}\\
        \midrule
        \multirow{3}{*}{PACS \& VLCS}       & learning rate & $5 \times 10^{-5}$ & - \\      
                                    & weight decay & 0 & - \\
                                      & $\gamma_1, \gamma_2$ & 0.1, 0.5 & - \\
        \midrule
        \multirow{3}{*}{OfficeHome}       & learning rate & $5 \times 10^{-5}$ & - \\      
                                      & weight decay & $1 \times 10^{-4}$ & - \\
                                      & $\gamma_1, \gamma_2$ & 0.5, 0.5 & - \\
        \midrule
        \multirow{3}{*}{TerraIncognita}       & learning rate & $2 \times 10^{-4}$ & - \\      
                                      & weight decay & $3 \times 10^{-6}$ & - \\
                                      & $\gamma_1, \gamma_2$ & 0.5, 0.5 & - \\
        \midrule
        \multirow{4}{*}{All above}   & steps & 5000 & - \\      
                                    & T & 1 & [0.5, 1, 2, 4, 8] \\
                                    & $N_{\textrm{anneal}}$ & 1000 & [100, 500, 1000, 2000] \\
                                    & $\lambda$ & 1 & [1, 5, 10, 50, 100, 200] \\
                                    & $N_{\textrm{warm-up}}$ & 100 & [50, 100, 200, 500, 1000, 2000]\\
        \midrule
        \multirow{8}{*}{DomainNet}       & learning rate & $2 \times 10^{-4}$ & - \\      
                                      & weight decay & $3 \times 10^{-6}$ & - \\
                                      & $\gamma_1, \gamma_2$ & 0.1, 0.5 & - \\
                                      & steps & 15000 & - \\      
                                    & T & 1 & [0.5, 1, 2, 4, 8, 16] \\
                                    & $N_{\textrm{anneal}}$ & 1000 & [1000, 2000, 5000, 10000] \\
                                    & $\lambda$ & 1 & [10, 50, 100, 200, 500, 1000] \\
                                    & $N_{\textrm{warm-up}}$ & 100 & [500, 1000, 2000, 5000, 10000]\\
        \midrule
        \multirow{2}{*}{All}       & dropout & 0 & - \\
                                    & batch size & 16 & - \\
        \bottomrule
    \end{tabular}
    }
    \end{center}
    \label{table:hyperparameters}
\end{table*}

\begin{table*}[ht]
\caption{Classification accuracy on TerraIncognita. Results denoted with $\dag$ are from ZooD~\cite{dong2022zood}.}
\label{aptab:terra}
\setlength\tabcolsep{5.2 pt}
\begin{center}
\begin{small}
\begin{tabular}{c|ccccc}
\toprule
{\multirow{2}{*}{\textbf{Method}}} & \multicolumn{5}{c|}{\textbf{Terra Incognita}} \\
& \textbf{L100}&\textbf{L38}&\textbf{L43}&\textbf{L46} & \textbf{Avg} \\
\midrule
Single$^\dag$ & 33.7 & 37.1 & 40.3 & 37.9 & 37.3 \\
Ensemble$^\dag$ & 35.2 & 34.1 & 45.6 & 37.9 & 38.2 \\
F. Selection$^\dag$ & 40.0 & 46.1 & 45.1 & 37.8 & 42.3  \\
\midrule

Single+rew & 58.9 +/- 2.7 & 50.5 +/- 0.3 & 53.5 +/- 0.4 & 45.1 +/- 3.1 & 52.0 +/- 1.3  \\
Single+hsic & 59.1 +/- 0.8 & 48.5 +/- 2.5 & 53.2 +/- 0.3 & 48.3 +/- 1.4 & 52.3 +/- 0.9  \\
Single+both & 61.7 +/- 1.4 & 47.6 +/- 2.0 & 53.0 +/- 0.9 & 49.5 +/- 0.5 & 53.0 +/- 0.9  \\
\bottomrule
\end{tabular}
\end{small}
\end{center}
\vskip -0.1in
\end{table*}

\begin{table*}[ht]
\caption{Classification accuracy on VLCS. Results denoted with $\dag$ are from ZooD~\cite{dong2022zood}.}
\label{aptab:vlcs}
\setlength\tabcolsep{5.2 pt}
\begin{center}
\begin{small}
\begin{tabular}{c|ccccc}
\toprule
{\multirow{2}{*}{\textbf{Method}}} & \multicolumn{5}{c}{\textbf{VLCS}} \\
 &\textbf{C}&\textbf{L}&\textbf{S}&\textbf{V} & \textbf{Avg} \\
\midrule
Single$^\dag$ & 99.9 & 60.5 & 72.3 & 85.4 & 79.5 \\
Ensemble$^\dag$ & 99.7 & 63.4 & 76.5 & 80.9 & 80.1 \\
F. Selection$^\dag$ & 100.0 & 63.0 & 77.0 & 82.4 & 80.6 \\
\midrule

Single+rew & 99.9 +/- 0.0 & 63.4 +/- 1.1 & 76.9 +/- 0.9 & 84.4 +/- 0.2 &  81.2 +/- 0.4 \\
Single+hsic & 99.7 +/- 0.1 & 65.7 +/- 0.5 & 75.4 +/- 0.4 & 85.2 +/- 0.1 & 81.5 +/- 0.1 \\
Single+both &  99.8 +/- 0.1 & 64.6 +/- 1.0 & 76.4 +/- 0.4 & 84.7 +/- 0.2 & 81.4 +/- 0.3 \\
\bottomrule
\end{tabular}
\end{small}
\end{center}
\vskip -0.1in
\end{table*}

\paragraph{Hyperparameters.} Following ZooD, we adopt the leave-one-domain-out cross-validation setup in DomainBed for hyper-parameter selection and run 3 trials. We list all hyperparameters, their default values, and the search range for each hyperparameter in our grid search sweeps, in Table \ref{table:hyperparameters}. All models are optimized using Adam~\citep{kingma2014adam}. Note that The hyperparameters $\gamma_1$ and $\gamma_2$ are intrinsic to each PTM and define the bandwidth of the Gaussian kernel in HSIC. They are manually set to ensure the penalty term's initial value falls within the proper range $[0, 1]$, which is influenced by the PTM's feature scale and range. Consequently, the value of $\gamma_1$ varies with the choice of the main model. For CLIP-ViT and ResNext-101, $\gamma_1$ is set to $0.1$, while for Swin-B, it is set to $0.5$. Empirically, we observe that results are not very sensitive to small deviations ($~0.25$) from these chosen values.

\subsection{Detailed Results}
We include some detailed results of the experiments on the DomainBed. Table~\ref{aptab:terra} shows the classification accuracy on Terra-Incognita. Table~\ref{aptab:vlcs} show the classification accuracy on VLCS. It shows that our proposed scheme outperforms ZooD on each target domain in the dataset.

\end{document}